%% file: DS-Det-arXiv.tex
\documentclass[sigconf]{acmart}

\usepackage{bbding}
\usepackage{multirow}
\usepackage{bm}

\usepackage{colortbl}
\definecolor{shapecolor}{rgb}{0.0,0.5,0.0}
\usepackage{subcaption}

\input{math_commands.tex}

\AtBeginDocument{%
  }

\copyrightyear{2025}
\acmYear{2025}
\setcopyright{acmlicensed}
\acmConference[MM '25]{Proceedings of the 33rd ACM International Conference on Multimedia}{October 27--31, 2025}{Dublin, Ireland}
\acmBooktitle{Proceedings of the 33rd ACM International Conference on Multimedia (MM '25), October 27--31, 2025, Dublin, Ireland}
\acmDOI{10.1145/3746027.3755045}
\acmISBN{979-8-4007-2035-2/2025/10}

\begin{document}

\title{DS-Det: Single-Query Paradigm and Attention Disentangled Learning for Flexible Object Detection}

\author{Guiping Cao}
\orcid{0000-0002-0682-2158}
\affiliation{%
  \institution{RITAS, Southern University of Science and Technology}
  \city{Shenzhen}
  \country{China}
}
\affiliation{%
  \institution{Pengcheng Laboratory}
  \city{Shenzhen}
  \country{China}
}
\email{12131099@mail.sustech.edu.cn}

\author{Xiangyuan Lan}
\authornote{Corresponding author.}
\orcid{0000-0001-8564-0346}
\affiliation{%
  \institution{Pengcheng Laboratory}
  \city{Shenzhen}
  \country{China}}
\affiliation{%
  \institution{Pazhou
Laboratory (Huangpu)}
  \city{Guangzhou}
  \country{China}}
\email{lanxy@pcl.ac.cn}

\author{Wenjian Huang}
\orcid{0000-0003-2408-8302}
\affiliation{%
  \institution{RITAS, Southern University of Science and Technology}
  \city{Shenzhen}
  \country{China}
}
\email{huangwj@sustech.edu.cn}

\author{Jianguo Zhang}
\authornotemark[1]
\orcid{0000-0001-9317-0268}
\affiliation{%
  \institution{RITAS and Department of Computer Science and Engineering, Southern University of Science and Technology}
  \city{Shenzhen}
  \country{China}
}
\affiliation{%
      \institution{Pengcheng Laboratory}
  \city{Shenzhen}
  \country{China}
}
\email{zhangjg@sustech.edu.cn}

\author{Dongmei Jiang}
\orcid{0000-0002-6238-8499}
\affiliation{%
  \institution{Pengcheng Laboratory}
  \city{Shenzhen}
  \country{China}}
\email{jiangdm@pcl.ac.cn}

\author{Yaowei Wang}
\orcid{0000-0002-6110-4036}
\affiliation{%
  \institution{Harbin Institute of Technology (Shenzhen)}
  \city{Shenzhen}
  \country{China}}
\affiliation{%
  \institution{Pengcheng Laboratory}
  \city{Shenzhen}
  \country{China}}
\email{wangyw@pcl.ac.cn}

\renewcommand{\shortauthors}{Cao et al.}

\input{sec/0_abstract}



\maketitle

\input{sec/1_intro}

\input{sec/2_related}

\input{sec/3_method}

\input{sec/4_experiments}

\input{sec/5_conclusion}

\section*{Acknowledgements}

This work was supported in part by the National Natural Science Foundation of China (Grant No. 62276121), in part by the National Natural Science Foundation of China (Grant No. 62402252), in part by the TianYuan funds for Mathematics of the National Science Foundation of China (Grant No. 12326604), in part by the Shenzhen International Research Cooperation Project (Grant No. GJHZ20220913142611021), and in part by the Pengcheng Laboratory Research Project.

\bibliographystyle{ACM-Reference-Format}
\balance
\bibliography{DS-Det-arXiv}

\input{sec/appendix}

\end{document}

%% file: math_commands.tex
\usepackage{amsmath,amsfonts,bm}

\def\eqref#1{equation~\ref{#1}}

\def\1{\bm{1}}

\def\vq{{\bm{q}}}

\def\mB{{\bm{B}}}

\def\mE{{\bm{E}}}

\DeclareMathAlphabet{\mathsfit}{\encodingdefault}{\sfdefault}{m}{sl}
\SetMathAlphabet{\mathsfit}{bold}{\encodingdefault}{\sfdefault}{bx}{n}



%% file: sec/0_abstract.tex
\begin{abstract}

Popular transformer detectors have achieved promising performance through \textit{query-based} learning using \textit{attention} mechanisms. However, the roles of existing decoder query types (\textit{e.g.}, content query and positional query) are still underexplored. These queries are generally predefined with a fixed number (\textit{fixed-query}), which limits their flexibility. We find that the learning of these fixed-query is impaired by \textit{Recurrent Opposing inTeractions} (ROT) between two attention operations: Self-Attention (query-to-query) and Cross-Attention (query-to-encoder), thereby degrading decoder efficiency. Furthermore, ``\textit{query ambiguity}'' arises when shared-weight decoder layers are processed with both one-to-one and one-to-many label assignments during training, violating DETR's one-to-one matching principle. To address these challenges, we propose \textbf{DS-Det}, a more efficient detector capable of detecting a flexible number of objects in images. Specifically, we reformulate and introduce a new unified \textit{Single-Query} paradigm for decoder modeling, transforming the fixed-query into flexible. Furthermore, we propose a simplified decoder framework through attention disentangled learning: locating boxes with Cross-Attention (one-to-many process), deduplicating predictions with Self-Attention (one-to-one process), addressing ``query ambiguity'' and ``ROT'' issues directly, and enhancing decoder efficiency. We further introduce a unified PoCoo loss that leverages box size priors to prioritize query learning on hard samples such as small objects. Extensive experiments across \textit{five} different backbone models on COCO2017 and WiderPerson datasets demonstrate the general effectiveness and superiority of DS-Det. The source codes are available at \url{https://github.com/Med-Process/DS-Det/}.

\end{abstract}

%% file: sec/1_intro.tex
\section{Introduction}
\label{sec:intro}

In the last few years, \textit{transformer} detectors like DETR~\cite{carion2020end} and DINO~\cite{zhang2022dino}, have simplified the detection pipeline by providing end-to-end detection capabilities through \textit{query-based} learning using attention mechanisms, demonstrating promising performance compared to classical CNN-based detectors~\cite{girshick2015fast, ren2015faster, liu2016ssd, redmon2016you, he2017mask}. Subsequently, a series of follow-up works have been proposed to boost DETR on the architecture of encoder and decoder~\cite{zhu2020deformable, cao2022cf}, query formulations~\cite{meng2021conditional, liu2022dab, caomlp_DINO}, and training efficacy~\cite{meng2021conditional, zhang2022accelerating, zhang2022dino, jia2023detrs, hu2024dac}.

\textbf{Query Components and Roles.}
DETR-like models~\cite{carion2020end, meng2021conditional, yao2021efficient, liu2022dab, li2022dn, zhang2022dino, zheng2023less, caomlp_DINO, cao2025cross} formulate object detection (OD) as a set prediction task using a transformer encoder-decoder architecture. These models detect objects through interactions in the decoder between \textit{object queries} and encoder features. Each query is responsible for the prediction of the class label and bounding box coordinates of a single object, following a one-to-one matching manner. However, \textit{the query components and roles are underexplored}~\cite{liu2022dab,zhang2022dino}. For example, the pioneering DETR~\cite{carion2020end} introduces two types of queries: one initialized as a zero vector and another with static learned embeddings, yet their semantic meanings for detection remain unclear. Although subsequent works~\cite{meng2021conditional,liu2022dab,li2022dn,zhang2022dino,hu2024dac} have further categorized these decoder queries into Content Query (CQ) and Positional Query (PQ), thus providing some explicit semantics, the precise roles and impacts of these queries for object detection remain understudied. This limitation constrains further performance enhancement for transformer detectors.

\textbf{Fixed-Query's Limitation.} Most existing DETR-like models~\cite{zhu2020deformable,zhang2022dino,liu2023detection,hu2024dac} predefine a \textit{fixed-size} set of $N$ learnable object queries (including both the PQ and CQ), which can only detect a fixed number of objects, thus limiting their flexibility in dense scenarios.

\textbf{Query Ambiguity.} Both CQ and PQ are updated layer-by-layer in the transformer decoder through Self-Attention (SA, query-to-query interactions) and Cross-Attention (CA, query-to-encoder feature interactions) mechanisms. However, existing researches~\cite{meng2021conditional,hu2024dac} demonstrate that these attention operations exert opposing effects: CA gathers multiple queries around the same object, while SA disperses queries from each other. These Recurrent Opposing inTeractions (ROT) in current decoder frameworks~\cite{carion2020end,li2022dn,zhang2022dino,hu2024dac,caomlp_DINO,huang2024dq} create ``\textit{query ambiguity}'', where queries struggle to both converge and diverge for the same object during the optimization process, thereby degrading decoder efficiency.

\begin{figure*}[t]
  \centering
   \includegraphics[width=0.77\linewidth]{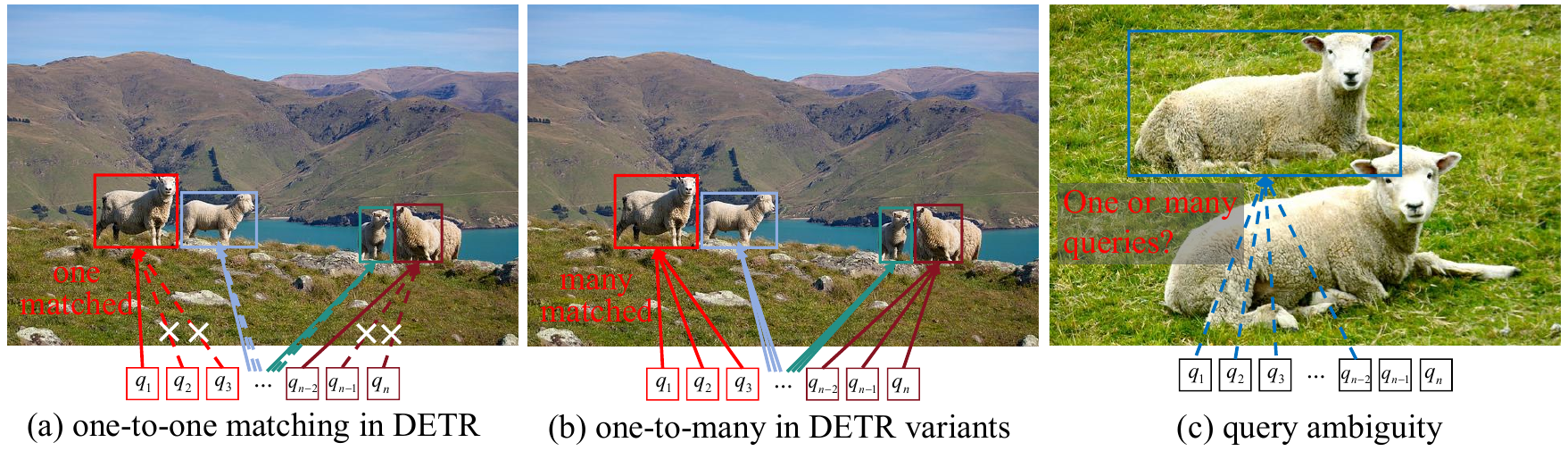}
   \caption{The mixing of one-to-one and one-to-many matching with shared weights in the decoder introduces \textit{query ambiguity}, causing multiple queries to predict the same single object and increasing false positives. 
   }
   \label{fig:one-to-one}
\end{figure*}

Furthermore, the mixing of \textit{one-to-one} matching (DETR's foundational property for end-to-end detection) and \textit{one-to-many} matching (the core method for improving training efficiency and convergence) also introduces ``\textit{query ambiguity}'' when applied to queries with shared decoder weights. Thus, models struggle to predict single or multiple results for the same object, as illustrated in Fig.~\ref{fig:one-to-one}(c).

\textbf{DS-Det.} To address these challenges, we first comprehensively evaluate the query components in transformer detectors, analyzing their functional roles. Building on this analysis, we reformulate the query composition mechanism by introducing a novel Single-Query paradigm that transforms traditional \textit{fixed-query} into \textit{flexible-query}. Furthermore, to address the ``\textit{query ambiguity}'' problem in decoder architectures, we systematically investigate CA and SA mechanisms and further develop a novel efficient decoder architecture that significantly improves its efficiency. Overall, our key contributions can be summarized as follows: 

(1) We propose an efficient transformer detector termed \textbf{DS-Det}, which reformulates the query paradigm through Flexible singLe-quEry generaTion (FLET) module, transforming the fixed-query into flexible-query and enhancing the model's detection flexibility.

(2) We design a novel \textit{Attention Disentangled Decoder} (ADD) architecture that disentangles the attentions of CA and SA and further separates one-to-one and one-to-many matching processes, efficiently addressing the issues of ``query ambiguity'' and ``ROT'' and significantly improving decoder efficiency.

(3) We introduce a new loss function, which incorPOrates Classification with iOu and bOx size (PoCoo) for re-weighting the classification loss, enhancing query's learning on hard samples.

(4) Extensive experiments on the COCO2017 and WiderPerson datasets demonstrate the general effectiveness of DS-Det across five backbone models. Furthermore, DS-Det shows significant advantages in flexible-query and decoder efficiency, improving the decoder inference speed by \textbf{+34.8\%}.

%% file: sec/2_related.tex
\section{Related Works}
\label{sec:formatting}

\quad \textbf{Transformer Detectors.} DETR~\cite{carion2020end}, as the pioneer transformer detector, represents a significant breakthrough in OD. By framing OD as a direct set prediction task and leveraging transformer architectures, DETR introduces a more efficient end-to-end detection paradigm, eliminating the need for hand-designed components like NMS. 
Since then, many follow-up efforts have focused on various aspects of DETR enhancement, including accelerating the model training ~\cite{meng2021conditional, li2022dn, zhang2022dino, jia2023detrs, zong2023detrs}, reformulating the decoder queries ~\cite{meng2021conditional, yao2021efficient, liu2022dab, zhang2023dense}, improving the encoder and decoder architectures ~\cite{zhu2020deformable, roh2021sparse, cao2022cf}, and optimizing loss functions ~\cite{liu2023detection, cai2023align, pu2024rank, hu2024dac}. 
While these methods effectively enhance detection capabilities, they often overlook the components and roles of decoder queries, as well as their fixed limitations in object prediction capacity.

\textbf{Query Paradigm.} Following the query paradigm of DETR~\cite{carion2020end}, many subsequent works~\cite{zhu2020deformable, meng2021conditional, yao2021efficient, wang2022anchor, liu2022dab, zhang2022dino} have made efforts to interpret queries more explicitly. Conditional DETR~\cite{meng2021conditional} categorizes these two query types as Content Query (CQ) and Positional Query (PQ), with the latter formulated as learnable 2D coordinates $(x, y)$. DAB-DETR~\cite{liu2022dab} further advances this formulation by employing 4D box coordinates $(x, y, w, h)$, thereby incorporating stronger spatial priors. To the best of our knowledge, almost all follow-up works~\cite{liu2023detection,hu2024dac,zhang2023dense,chen2023group,caomlp_DINO,huang2024dq} have adopted these two query types, mainly differing in their initialization strategies. However, the specific roles of these queries (particularly for the PQ) in the decoder's detection process remain unexplored. This knowledge gap may substantially compromise detection performance.

\textbf{Query Updates.} Most transformer detectors~\cite{carion2020end,zhu2020deformable,li2022dn,zhang2022dino,zhang2023dense,hu2024dac,caomlp_DINO} follow a standard decoder architecture where queries interact by two attention mechanisms: self-attention (SA) for query-to-query relations and cross-attention (CA) for query-to-encoder updates. However, DAC-DETR~\cite{hu2024dac} reveals that these operations exert opposing effects on queries: CA gathers multiple queries around the same object, while SA disperses them. In this paper, we explore and identify this fundamental conflict as ``Recurrent Opposing inTeractions'' (ROT) issue, systematically degrading decoder efficiency.

\textbf{Query Matching.} DETR~\cite{carion2020end} and its variants~\cite{zhu2020deformable,liu2022dab,zhang2022dino}, innovate with a \textbf{one-to-one matching} approach for end-to-end object detection, as shown in Fig.~\ref{fig:one-to-one} (a), eliminating the need for post-processing to remove duplicate detections. Though streamlining the detection workflow, this one-to-one matching manner leads to only a few queries assigned as positive samples, thereby significantly diminishing the training efficiency of positive samples due to sparse supervision. To address this, many efforts, including Hybrid-DETR~\cite{jia2023detrs}, Co-DETR~\cite{zong2023detrs}, Group-DETR~\cite{chen2023group}, Align-DETR~\cite{cai2023align} and DAC-DETR~\cite{hu2024dac}, etc, have explored \textbf{one-to-many} label assignments for increasing the matched positive samples among dense queries.
By explicitly assigning multiple queries to each ground truth box, these methods boost the quantity of positive matches, accelerate model convergence, and enhance training efficiency.

However, the shift to one-to-many matching naturally introduces \textit{``query ambiguity''}, conflicting with DETR's one-to-one principle. This ambiguity arises from mixing one-to-many and one-to-one matching via shared weights in auxiliary decoders or branches, leading to uncertainty in predicting single or multiple results per object, as depicted in Fig.~\ref{fig:one-to-one} (c).
Unfortunately, this issue is often overlooked in existing works~\cite{carion2020end, zhang2022dino, jia2023detrs, cai2023align, hu2024dac}. Furthermore, the reliance on \textit{additional} decoder branches for one-to-many matching (\textit{e.g.}, Hybrid-DETR~\cite{jia2023detrs} and DAC-DETR~\cite{hu2024dac}) increases training complexity and computational costs, worsening the ambiguity.

\textbf{Flexible Object Detection.} In real-world scenarios, the number of detectable objects varies widely~\cite{shao2018crowdhuman, zhang2019widerperson}, ranging from individual instances to thousands, presenting significant challenges for detectors~\cite{liu2021survey, cheng2023towards}. 
Recently, DiffusionDet~\cite{chen2023diffusiondet} introduced a novel framework that formulates OD as a diffusion denoising process from numerous noisy boxes to refined object boxes, 
achieving the flexibility to predict a variable number of detections by decoupling training and evaluation processes and leveraging iterative evaluation.
However, despite its advancements, DiffusionDet still suffers from several limitations. Notably, it requires hand-designed post-processing with NMS for duplicate box removal, complicating both the training and inference process; additionally, its adaptability is constrained by \textit{manually defined parameters of noisy box number} and evaluation iterations, increasing the evaluation complexity; moreover, DiffusionDet still lags behind the SOTA works such as DINO~\cite{zhang2022dino}, hindering its potential for development and application. 
Recently, DQ-DETR~\cite{huang2024dq} introduces a categorical counting module to address tiny object detection by discretizing query numbers into fixed bins (300/500/900/1500). However, this classification approach inherently lacks the flexibility to dynamically determine image-specific optimal query counts.

%% file: sec/3_method.tex
\section{Methods}

We first analyze the limitations of DETR-like models by rethinking the roles of queries (CA and PQ) and attention mechanisms. Next, we present the overall architecture of DS-Det and propose the novel FLET modules to enable ``flexible predictions''.
We then introduce our Attention Disentangled Decoder (ADD) approach to address the ``query ambiguity'' and ``ROT'' issues.
Finally, we propose the PoCoo loss for enhancing the query's learning on hard samples.

\subsection{Roles of Queries and Attention Mechanisms}
\label{sec:role_query_attention}

\textbf{Revisiting Content and Positional Queries.} 
CQ and PQ are fundamental components to existing DETR-like frameworks. While \textbf{CQ}, as decoder embeddings in DETR, is directly used for predicting boxes and classes, \textbf{PQ}'s role in enhancing query diversity~\cite{carion2020end} remains unclear.
Despite optimizations in prior works~\cite{zhu2020deformable,meng2021conditional,yao2021efficient,wang2022anchor,liu2022dab,zhang2022dino}, further investigation is needed to clarify the role of PQ in detection. Consequently, we conduct ablation experiments with DINO~\cite{zhang2022dino} on COCO~\cite{lin2014microsoft} by removing PQ to evaluate its impact. Results in Table~\ref{tab:pos_query} show that PQ removal has a minimal effect, with average precision (AP) varying by only $\pm{\textbf{0.3\%}}$.
To transfer fixed-query into flexible-query, we eliminate the fixed-number PQ and CQ and further introduce a Single-Query paradigm (in Sec.~\ref{sec:flet}), simplifying query's components and improving its efficiency.

\begin{table}[h]
\centering
     \caption{The ablations on the PQ in SA and CA.
     }
\resizebox{1.0\columnwidth}{!}{
\setlength{\tabcolsep}{2.0mm}{
  \renewcommand\arraystretch{0.1}
\begin{tabular}{ c c c c }
    \toprule
     Model & PQ in SA & PQ in CA & AP \\
     \midrule
     \multirow{4}{*}[-2.0ex]{DINO-ResNet50~\cite{zhang2022dino}} & \checkmark & \checkmark & 49.5 \\
     ~ & \checkmark & \XSolidBrush & 49.8 (+0.3)  \\
     ~ & \XSolidBrush & \checkmark & 49.3 (-0.2)  \\
     ~ & \XSolidBrush & \XSolidBrush & 49.3 (-0.2) \\
     \midrule
     \multirow{4}{*}[-2.0ex]{DINO-Strip-MLP-T~\cite{zhang2022dino}} & \checkmark & \checkmark & 51.7 \\
     ~ & \checkmark & \XSolidBrush & 51.6 (-0.1) \\
     ~ & \XSolidBrush & \checkmark & 51.8 (+0.1) \\
     ~ & \XSolidBrush & \XSolidBrush & 51.7 \\
     \bottomrule
     \end{tabular}}
     }
\label{tab:pos_query}
\end{table}

\textbf{High GPU Memory Demand in SA.} The query-to-query interactions in SA pose training challenges for flexible detection, as their computational complexity scales with the number of queries increasing. 
Specifically, SA processes $N$ queries $Q=\{q_1, ..., q_n\}$ with $O(n^{2})$ complexity from the $QK^{T}$ operation ( $Q$ and $K$ are query and key). For flexible detection training, query numbers may scale up (\textit{e.g.}, $N=20,000$), resulting in prohibitive GPU memory usage and \textit{infeasible training}. This raises critical questions: What is the role of SA? Can it be removed to address GPU Memory issues?

\textbf{Role and Necessity of SA in Decoder.} To answer these two questions, we conduct SA ablations with DINO on COCO in Table~\ref{tab:sa_linear}. Removing SA causes a \textbf{4.9\%} AP drop, but only a \textbf{0.6\%} average recall (AR) decrease. It suggests that \textit{SA significantly reduces false positives}. To transform the fixed-query into ``flexible-query'', the natural idea is to introduce a binary classification branch for the encoder to dynamically select queries based on binary prediction results (positive and negative samples). We achieve this model (noted ``Binary'') and yields a \textbf{+2.7\%} AP gain over the baseline, and adding NMS further improves AP by \textbf{+0.7\%}, demonstrating SA's performance boost is comparable to NMS.

\vspace{-1ex}
\begin{table}[h]
\small
\centering
     \caption{Ablations on SA layers. 
     }
\vspace{-1ex}
\resizebox{1.0\columnwidth}{!}{
\setlength{\tabcolsep}{0.5mm}{
 \renewcommand\arraystretch{1.0}
\begin{tabular}{ c  c  c  c c c  c c c }
    \toprule
     Method & SA & Query Number & AP &  AP$_{S}$ & AP$_{L}$ & AR & AR$_{S}$ & AR$_{L}$ \\
     \midrule

    \multirow{2}{*}[-0.5ex]{Baseline} & \checkmark & fixed & 49.0 & 32.0 & 63.0 & 72.7 & 55.9 & 88.4 \\
     ~ & \XSolidBrush & fixed & 44.1(-4.9) & 27.9 & 56.7 & 72.1(-0.6) & 53.3 & 86.6 \\
    \midrule
    Binary & \XSolidBrush & \textit{flexible} & 46.8 & 29.9 & 59.7 & 71.9 & 53.6 & 87.5 \\
    Binary-NMS & \XSolidBrush & \textit{flexible} & 47.5 & 30.5 & 61.3 & 69.3 & 53.9 & 83.3 \\
    \midrule
    Binary-Linear-Attn & \checkmark & \textit{flexible} & 48.4 & 30.4 & 62.4 & 73.9 & 58.9 & 88.0 \\ 
    Binary-SA & \checkmark & \textit{flexible} & 49.1 & 31.4 & 63.4 & 73.1 & 57.9 & 88.3 \\
    
     \bottomrule
     \end{tabular}}
     }
\label{tab:sa_linear}
\end{table}

To mitigate the quadratic complexity of SA in standard transformer~\cite{vaswani2017attention}, we further construct a new variant of Binary-Linear-Attn, by employing \textbf{linear transformer} architectures with a binary head. These linear transformers, such as efficient attention~\cite{shen2021efficient} and external attention~\cite{guo2023beyond}, eliminate \textit{pairwise} attention score computations between all input tokens, reducing complexity. 
However, experiments in Table~\ref{tab:sa_linear} show that Binary-SA outperforms Binary-Linear-Attn, indicating that \textit{computing attention maps between all query pairs is more efficient in eliminating duplicates}.
These findings emphasize SA's critical role in eliminating duplicates, motivating our DP design in ADD (Sec.~\ref{sec:ldd}) that leverages SA to eliminate duplicates by computing similarities between query pairs and refining queries through one-to-one matching and class supervision.

\textbf{Opposing Roles of CA and SA.} 
Most DETR-like \textit{one-to-many} decoder architectures~\cite{jia2023detrs, zong2023detrs, hu2024dac} struggle with \textit{query ambiguity} due to two main reasons: (1) the mixing of one-to-one and one-to-many label matching, and (2) the opposing effects of CA and SA on object queries.
SA disperses queries from each other, while CA clusters them around the same object~\cite{hu2024dac}. 
In our analysis, CA transfers object information from encoder features into decoder queries, linking multiple queries to the same object to enhance bounding box accuracy (the one-to-many process). However, this process inevitably increases false positives.
Most existing decoder architectures apply the one-to-one process of SA before the one-to-many process of CA layer-by-layer. This \textbf{\textit{recurrent opposing interactions}} between these two opposing processes leads to a significant problem of ``query ambiguity''.

\begin{figure*}[!t]
   \centering
    \includegraphics[width=0.85\linewidth]{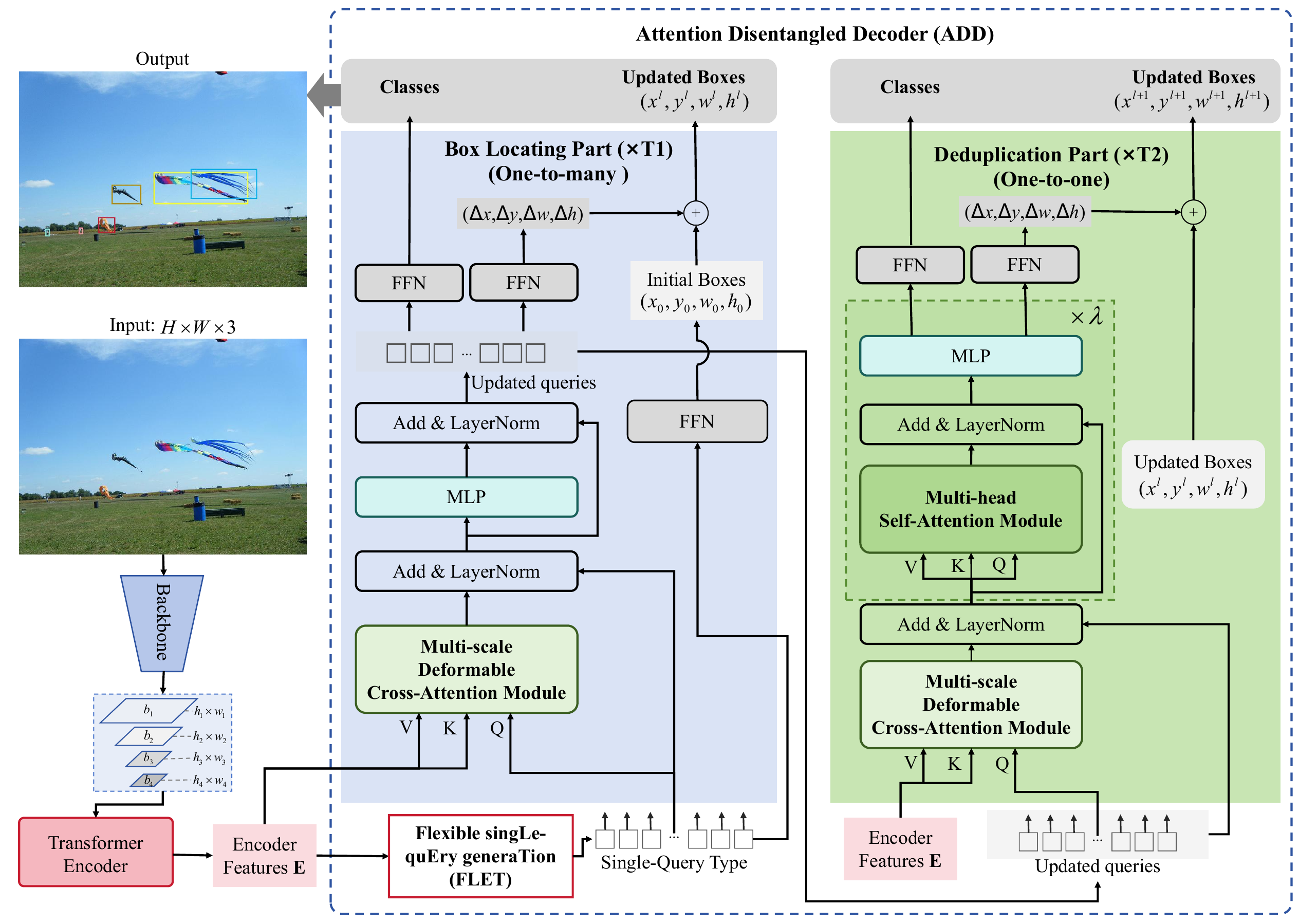}
    \caption{The overall architecture of the DS-Det model.
    }
    \label{fig:qfree-det}
 \end{figure*}

Based on the above analysis, we propose a new ADD framework in Sec.~\ref{sec:ldd}, which disentangles the CA and SA into two simple parts: box locating with CA and duplicate detections removing with SA. ADD mitigates the query ambiguity while \textit{retaining the benefits of CA and SA}, leading to a more efficient decoder architecture.

\subsection{Main Architecture of DS-Det}

Fig.~\ref{fig:qfree-det} illustrates an overview of DS-Det model. The input image $\bm{I} \in \mathbb{R}^{H \times W \times 3}$ is processed by the backbone and transformer encoder to obtain the enhanced feature representations $\bm{E}$. 
Then, the proposed FLET module operates on all encoder tokens to generate a flexible number of decoder queries. These queries, along with the encoder features $\bm{E}$, are fed into the \textbf{Box Locating Part} (BLP) for $T_1$ iterations, aiming to locate the bounding box of each object using multiple queries and maintain training efficacy through one-to-many label matching. Next, the \textbf{Deduplication Part} (DP), consisting of $T_2$ iterations, takes $\bm{E}$ and the updated decoder queries as input to remove duplicate detections via one-to-one matching.
Each decoder layer outputs bounding boxes and classification results for detected objects. The initial bounding box locations $\mB_{x_{0}y_{0}w_{0}h_{0}}$ are predicted by the encoder and refined layer-by-layer in the decoder through box offset regression. Mathematically, this process can be represented as:
\begin{equation}
    \mB_{x_{0}y_{0}w_{0}h_{0}} = FFN(FLET(\mE)) 
\end{equation}
\begin{equation}
    \mB_{xywh}^{i} = \mB_{x_{0}y_{0}w_{0}h_{0}} + \sum \limits_{i=1}^{T_1}{BLP(\vq_d^{i}, \mE)} + \sum \limits_{\substack{j=i - T_1\\i > T_1}}^{T_2}{DP(\vq_d^{j}, \mE)}
\end{equation}
where $\vq_d$ denotes the feature of the Single-Query type; $i$ and $j$ indexes the layer of $\vq_d$.

\subsection{Flexible Single-Query Paradigm}
\label{sec:flet}

\begin{figure}[h]
   \centering
    \includegraphics[width=0.76\linewidth]{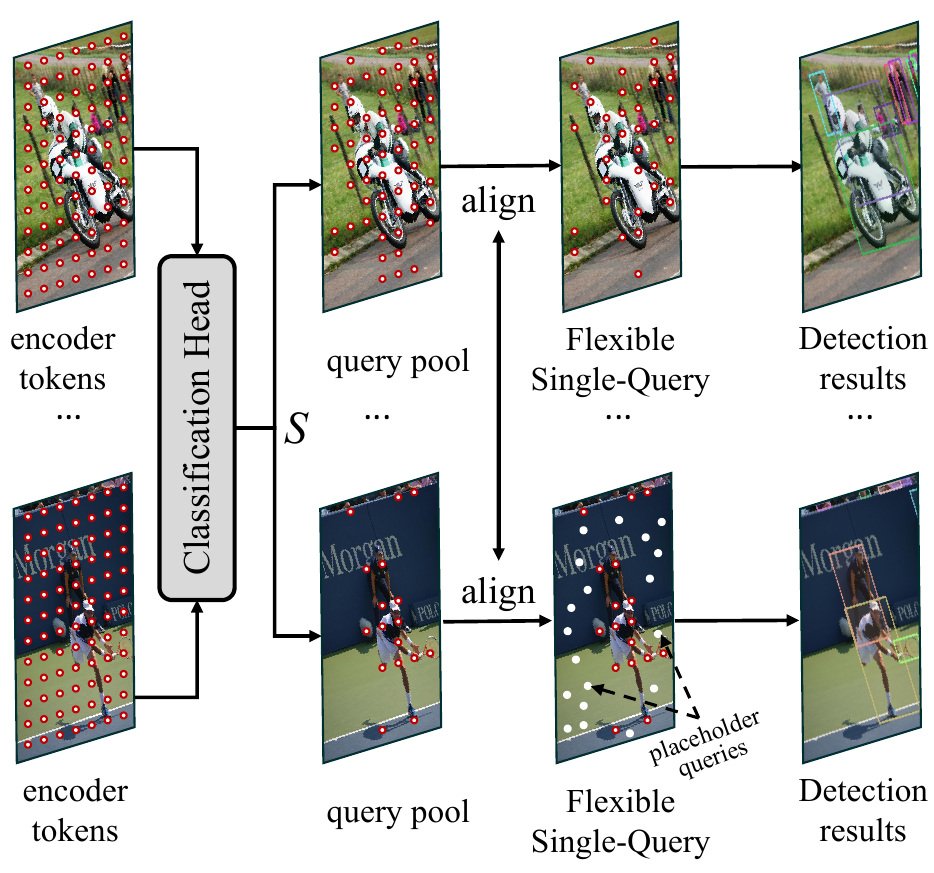}
    \caption{The main pipeline of the FLET module.
    }
    \label{fig:flet}
 \end{figure}

\quad \textbf{Single-Query Paradigm.} As illustrated in Fig.~\ref{fig:flet}, the FLET module is designed to simplify the query components and transfer the fixed-query into the flexible-query. Our method generates Single-Query through a novel encoder-to-query mechanism: (1) All encoder tokens are processed by a classification head to establish a global representation space; (2) A threshold $S$ selects the most informative tokens as queries, forming an adaptive query pool. Unlike traditional approaches that rely on static initialization~\cite{carion2020end,li2022dn} or anchor box embeddings~\cite{meng2021conditional,liu2022dab,zhang2022dino}, our method directly utilizes encoder tokens as queries, preserving richer object information and unifying CQ and PQ into the novel Single-Query paradigm.

\textbf{Query Alignment.} Maintaining a consistent number of queries per image within a batch is crucial for practical training. However, query pool sizes naturally vary across different images. To address this, we propose a Query Alignment method that operates as follows: For a batch size $b$, let $N_{query}=\{n_1, ..., n_b\}$ denote the number of queries for each image. We then define the batch query number $N_{query}^{b}$ as $max(N_{query})$, ensuring sufficient queries for all batch images. For images where FLET generates queries fewer than $N_{query}^b$, we introduce \textit{placeholder queries} by: (1) sorting all encoder tokens by ascending classification scores, and (2) selecting tokens with the lowest scores to minimize similarity with active queries.

\textbf{Query Sampling.} During \textit{early training stages}, classification scores of encoder tokens exhibit random distributions, leading to numerous redundant queries that significantly increase GPU memory demands for SA computations (analyzed in Sec.~\ref{sec:role_query_attention}). To mitigate this, we implement query sampling from the query pool, which simultaneously enables effective training and duplicate queries removal. As training progresses, classification scores demonstrate polarization: selected queries show progressively higher scores while others decrease, thereby naturally reducing query redundancy. This adaptive mechanism maintains query flexibility for diverse image detection scenarios.

\subsection{Attention Disentangled Decoder}
\label{sec:ldd}

To effectively mitigate ``\textit{query ambiguity}'' while maintaining training efficiency, we decompose the detection pipeline into two distinct stages: (1) localization (Box Locating Part) and (2) recognition with duplicate suppression (Deduplication Part), implemented through our proposed Attention Disentangled Decoder (ADD) architecture. ADD separates the attention mechanisms by: applying cross-attention with one-to-many matching for localization and employing self-attention with one-to-one matching for recognition.

\textbf{Box Locating Part (BLP).} The BLP is specifically designed to locate potential objects accurately through a one-to-many matching mechanism using multiple queries. This is achieved by replicating each ground truth box $K$ times and aligning each replicated box with a unique query. Given the distinct roles of CA in gathering multiple queries around the same object and SA in dispersing them, only CA is utilized in this stage for object prediction. Mathematically, this process can be formulated as:
\begin{equation}
\begin{split}
    \text{BLP}(\vq_d, \mE) = &LN\{MLP(LN(\vq_d + CA(\vq_d, \mE))) + \\
    &LN(\vq_d + CA(\vq_d, \mE))\}
\end{split}
\end{equation}
where LN is the LayerNorm~\cite{ba2016layer} layer. The MLP comprises two linear layers to enable channel-wise information interaction. $CA$ is the cross-attention operation.

\textbf{Deduplication Part (DP).} One-to-many matching mechanism inevitably introduces numerous false positives. To eliminate these duplicate queries, we further introduce the Deduplication Part (DP), which performs one-to-one matching to remove deplicates. 
The DP consists of two key components: the DP$_{\text{CA}}$ module, which refines the box location using the CA with the one-to-one matching scheme; the multi-head self-attention block (MSAB) that integrates both SA and MLP layers to eliminate duplicate detections. By incorporating the SA module and one-to-one label alignment, the DP achieves more effective one-to-one matching process. This deduplication process is iterated $\lambda$ times for enhancing performance, as formally described below:
\begin{equation}
    \text{DP}_{\text{CA}}(\vq_d, \mE) = LN(\vq_d + CA(\vq_d, \mE))
\end{equation}
\begin{equation}
\begin{split}
    \text{DP}_{\text{MSAB}}(\vq_d, \mE) = &\sum_{m=1}^{\lambda} MLP^m\{ LN^m\left[\vq_d^{m-1} + \right.\\
    &\left.SA^m(\vq_d^{m-1})\right]\}
\end{split}
\end{equation}
where $\lambda$ indexes the block number. $SA$ denotes the multi-head self-attention module~\cite{vaswani2017attention}. The $\vq_d^{0}$ is obtained from $\text{DP}_{\text{CA}}$, and $\vq_d^{m}$ is updated by $\text{DP}_{\text{MSAB}}$.
The MSAB block significantly reduces false positives by leveraging multiple SA layers. During training, the pairwise attention scores between all queries ensure that the assigned query for each ground truth object gradually updates to a higher classification score. In contrast, redundant queries associated with the same ground truth receive lower scores through one-to-one matching and classification supervision.

\textbf{Stop Gradient from DP to BLP.} 
Since the BLP and DP modules sequentially update queries, the one-to-many matched queries in BLP may later be matched in a one-to-one fashion in DP. This can result in \textit{conflicting supervision} and unstable gradient updates, reintroducing detection ambiguity.
To resolve this, we introduce \textbf{S}topping \textbf{G}radient back-propagation of \textbf{Q}ueries (SGQ) from DP to BLP during training. This ensures consistent matching across both parts, as demonstrated in Table~\ref{tab:sgq}.

\begin{table*}[!th]
    \caption{Comparison with popular detectors on \texttt{val2017} of COCO. The FLOPs of DS-Det are calculated on a 1280$\times$800 resolution with 900 queries, which matches the configuration of the baseline DINO. ``Classified'' refers to the model (DQ-DETR~\citep{huang2024dq}) predicting the number of queries among the predefined fixed bins (300/500/900/1500).
    } 
\centering
\resizebox{2.0\columnwidth}{!}{
\setlength{\tabcolsep}{1.0mm}{
 \renewcommand\arraystretch{0.7}
\begin{tabular}{ c c c c c  c c c c c c c c } 
    \toprule
     Model & Year & Backbone & Objects & Epochs & AP & AP$_{50}$ & AP$_{75}$ & AP$_S$ & AP$_{M}$ & AP$_{L}$ & Params & FLOPs \\
     \midrule
     DETR~\citep{carion2020end} & 2020 & ResNet50 & fixed & 500 & 42.0 & 62.4 & 44.2 & 20.5 & 45.8 & 61.1 & 41M & 86G \\
    Deformable-DETR~\citep{zhu2020deformable} & 2020 & ResNet50 & fixed & 50 & 46.2 & 65.2 & 50.0 & 28.8 & 49.2 & 61.7 & 40M & 173G \\
    Conditional DETR~\citep{meng2021conditional} & 2021 & ResNet50 & fixed & 108 & 43.0 & 64.0 & 45.7 & 22.7 & 46.7 & 61.5 & 44M & 90G \\
    Sparse-DETR~\citep{roh2021sparse} & 2021 & ResNet50 & fixed & 50 & 46.3 & 66.0 & 50.1 & 29.0 & 49.5 & 60.8 & 41M & 136G \\
    DAB-DETR~\citep{liu2022dab} & 2022 & ResNet50 & fixed & 50 & 42.6 & 63.2 & 45.6 & 21.8 & 46.2 & 61.1 & 44M & 100G \\
    Efficient-DETR~\citep{yao2021efficient} & 2021 & ResNet50 & fixed & \cellcolor{gray!80}36 & 44.2 & 62.2 & 48.0 & 28.4 & 47.5 & 56.6 & 32M & 159G \\
    CF-DETR~\citep{cao2022cf} & 2022 & ResNet50 & fixed & \cellcolor{gray!80}36 & 47.8 & 66.5 & 52.4 & 31.2 & 50.6 & 62.8 & - & -\\
    Focus-DETR~\citep{zheng2023less} & 2023 & ResNet50 & fixed & \cellcolor{gray!80}36 & 50.4 & 68.5 & 55.0 & 34.0 & 53.5 & 64.4 & 48M & 154G \\

    \midrule
    DiffusionDet~\citep{chen2023diffusiondet} & 2023 & ResNet50 & \textit{flexible} & \cellcolor{gray!100}60 & 46.8 & 65.3 & 51.8 & 29.6 & 49.3 & 62.2 & - & -\\
    
    \midrule
    Grounding DINO~\citep{liu2023grounding} & 2023 & ResNet50 & fixed & \cellcolor{gray!20}12 & 48.1 & 65.8 & 52.3 & 30.4 & 51.3 & 62.3 & - & - \\
    Co-DETR-4scale~\cite{zong2023detrs} & 2023 & ResNet50 & fixed & \cellcolor{gray!20}12 & 49.5 & 67.6 & 54.3 & 32.4 & 52.7 & 63.7 & - & - \\
    Co-DETR-5scale~\cite{zong2023detrs} & 2023 & ResNet50 & fixed & \cellcolor{gray!20}12 & 52.1 & 69.4 & 57.1 & 35.4 & 55.4 & 65.9 & - & 860G \\
    Stable-DINO~\cite{liu2023detection} & 2023 & ResNet50 & fixed & \cellcolor{gray!20}12 & 50.4 & 67.4 & 55.0 & 32.9 & 54.0 & 65.5 & 47M & 279G  \\
    StageInteractor~\cite{teng2023stageinteractor} & 2023 &	ResNet50 & fixed & \cellcolor{gray!20}12 & 46.3	& 64.3 & 50.6	& 29.8	& 49.6	& 60.8 & - & - \\ 
    DEQDet~\cite{wang2023deep} & 2023 & ResNet50 & 	fixed & \cellcolor{gray!20}12 &	46.6 &	65.3 &	50.6 & 30.5 & 	49.4 &	61.2 & 65M & - \\
    DETA~\cite{ouyang2022nms} & 2022 & ResNet50 & fixed & \cellcolor{gray!20}12 & 50.5 & 67.6 & 55.3 & 33.1 & 54.7 & 65.2 & 52M & - \\
    DDQ DETR~\cite{zhang2023dense} & 2023 & ResNet50 & fixed & \cellcolor{gray!20}12 & 51.3 & 68.6 & 56.4 & 33.5 & 54.9 & 65.9 & - & - \\
    MS-DETR~\cite{zhao2024ms-detr} & 2024 & ResNet50 & fixed & \cellcolor{gray!20}12 & 50.3 & 67.4 & 55.1 & 32.7 & 54.0 & 64.6 & - & - \\
    Align-DETR~\cite{cai2023align} & 2023 & ResNet50 & fixed & \cellcolor{gray!20}12 & 50.2 & 67.8 & 54.4 & 32.9 & 53.3 & 65.0 & 47M & 279G \\
    DAC-DETR~\cite{hu2024dac} & 2024 & ResNet50 & fixed & \cellcolor{gray!20}12 & 50.0 & 67.6 & 54.7 & 32.9 & 53.1 & 64.2 & - & -\\
    DINO~\citep{zhang2022dino} & 2022 & ResNet50 & fixed & \cellcolor{gray!20}12 & 49.0 & 66.6 & 53.5 & 32.0 & 52.3 & 63.0 & 47M & 279G  \\
    \textbf{DS-Det (ours)} & 2025 & ResNet50 & \textit{flexible} & \cellcolor{gray!20}12 & 50.5 +(1.5) & 67.5 & 55.1 & \textbf{34.3 (+2.3)} & 54.6 & 64.5 & 48M & 275G\\
    \midrule

    Align-DETR~\cite{cai2023align} & 2023 & ResNet50 & fixed & \cellcolor{gray!50}24 & 51.3 & 68.2 & 56.1 & 35.5 & 55.1 & 65.6 & 47M & 279G \\
    DAC-DETR~\cite{hu2024dac} & 2024 & ResNet50 & fixed & \cellcolor{gray!50}24 & 51.2 & 68.9 & 56.0 & 34.0 & 54.6 & 65.4 & - & - \\
    DQ-DETR~\citep{huang2024dq} & 2024 & ResNet50 & classified & \cellcolor{gray!50}24 & 50.2 & 67.1 & 55.0 & 31.9 & 53.2 & 64.7 & - & -\\
    DINO~\citep{zhang2022dino} & 2022 & ResNet50 & fixed & \cellcolor{gray!50}24 & 50.4 & 68.3 & 54.8 & 33.3 & 53.7 & 64.8 & 47M & 279G  \\
    DINO~\citep{zhang2022dino} & 2022 & ResNet50 & fixed & \cellcolor{gray!80}36 & 50.9 & 69.0 & 55.3 & 34.6 & 54.1 & 64.6 & 47M & 279G \\
    \textbf{DS-Det (ours)} & 2025 & ResNet50 & \textit{flexible} & \cellcolor{gray!50}24 & \textbf{51.3 (+0.9)} & 68.4 & 55.9 & \textbf{35.5 (+2.2)} & 54.8 & 65.7 & 48M & 275G\\
    \midrule

    DINO~\cite{zhang2022dino} & 2022 & Strip-MLP-T & fixed & \cellcolor{gray!20}12 & 51.7 & 69.5 & 56.8 & 34.7 & 55.1 & 66.0 & 44M & 263G \\
    \textbf{DS-Det (ours)} & 2025 & Strip-MLP-T & \textit{flexible} & \cellcolor{gray!20}12 & 52.8 (+1.1) & 70.1 & 57.6 & 35.7 (+1.0) & 56.8 & 67.4 & 45M & 264G \\
    \textbf{DS-Det (ours)} & 2025 & Strip-MLP-T & \textit{flexible} & \cellcolor{gray!50}24 & 54.5 & 72.0 & 59.5 & 37.4 & 58.5 & 69.6 & 45M & 264G \\
    \textbf{DS-Det (ours)} & 2025 & Strip-MLP-T & \textit{flexible} & \cellcolor{gray!80}36 & \textbf{55.0} & \textbf{72.5} & \textbf{60.0} & \textbf{38.4} & \textbf{58.9} & \textbf{69.9} & 45M & 264G  \\
    \midrule

    $\mathcal{H}$-Deformable-DETR~\citep{jia2023detrs} & 2023 & Swin-T & fixed & \cellcolor{gray!20}12 & 50.6 & 68.9 & 55.1 & 33.4 & 53.7 & 65.9 & - & - \\
    $\mathcal{H}$-Deformable-DETR~\citep{jia2023detrs} & 2023 & Swin-T & fixed & \cellcolor{gray!80}36 & 53.2 & 71.5 & 58.2 & 35.9 & 56.4 & 68.2 & - & - \\
    DINO~\citep{ren2023detrex} & 2023 & Swin-T & fixed & \cellcolor{gray!20}12 & 51.3 & 69.0 & 56.0 & 34.5 & 54.4 & 66.0 & 48M & 280G  \\
    StageInteractor~\cite{teng2023stageinteractor} & 2023 & Swin-S	& fixed & \cellcolor{gray!80}36	& 52.7	& 71.7 &	57.7 &	36.1 &	56.2 &	67.7 & - & - \\
    DEQDet~\cite{wang2023deep} & 2023 &	Swin-S	& fixed & \cellcolor{gray!50}24	& 52.7	& 72.3	& 57.6	& 36.6	& 55.9	& 68.4 & 90M & - \\
    \textbf{DS-Det (ours)} & 2025 & Swin-T & \textit{flexible} & \cellcolor{gray!20}12 & 52.7 (+1.4) & 70.2 & 57.6 & 36.3 (+1.8) & 56.4 & 67.7 & 49M & 281G \\
    \textbf{DS-Det (ours)} & 2025 & Swin-T & \textit{flexible} & \cellcolor{gray!50}24 & 54.4 & 71.9 & 59.3 & \textbf{39.1} & 58.1 & 69.2 & 49M & 281G \\
    \textbf{DS-Det (ours)} & 2025 & Swin-T & \textit{flexible} & \cellcolor{gray!80}36 & \textbf{54.9} & \textbf{72.4} & \textbf{59.9} & 38.3 & \textbf{58.6} & \textbf{69.6} & 49M & 281G \\

    \midrule
    DINO~\cite{zhang2022dino} & 2022 & Swin-L & fixed & \cellcolor{gray!20}12 & 56.8 & 75.6 & 62.0 & 40.0 & 60.5 & 73.2 & 218M & 945G \\
    \textbf{DS-Det (ours)} & 2025 & Swin-L & \textit{flexible} & \cellcolor{gray!20}12 & \textbf{57.7 (+0.9)} & \textbf{75.6} & \textbf{63.0} & \textbf{40.0} & \textbf{62.5} & \textbf{74.2} & 219M & 946G \\

    \midrule
    DINO~\citep{zhang2022dino} & 2022 & Swin-L & fixed & \cellcolor{gray!80}36 & 58.0 & 76.1 & \textbf{64.0} & 40.1 & 62.2 & \textbf{74.3} & 218M & 945G \\
    \textbf{DS-Det (ours)} & 2025 & Swin-L & \textit{flexible} & \cellcolor{gray!50}\textbf{24} & \textbf{58.2} & \textbf{76.1} & 63.6 & \textbf{41.6} & \textbf{62.7} & 74.2 & 219M & 946G \\

    \midrule
    
    DINO & 2024 & VMamba-T & fixed & \cellcolor{gray!20}12 & 53.5 & 71.5 & 58.2 & 36.6 & 56.7 & 68.3 & 50M & 290G  \\
    \textbf{DS-Det (ours)} & 2025 & VMamba-T & \textit{flexible} & \cellcolor{gray!20}{12} & \textbf{54.4 (+0.9)} & \textbf{72.0} & \textbf{59.2} & \textbf{38.8 (+2.2)} & \textbf{58.2} & \textbf{69.2} & 51M & 290G\\ 
    \bottomrule
     \end{tabular}}
}
\label{tab:sota}
\end{table*}

\subsection{Classification Loss with IoU and Box Size} 

To address the misalignment of queries between classification scores and box regression results, Align-DETR~\cite{cai2023align} introduced the IA-BCE loss, which combines IoU and classification scores as a new label $t$ in binary cross entropy (BCE) loss. This aligns the two scores effectively.
To further improve the query learning on hard samples like small objects, we develop a new unified PoCoo loss that incor\textbf{PO}rates \textbf{C}lassification with I\textbf{O}U and B\textbf{O}x Size:
\begin{equation}
\begin{split}
    \text{PoCoo} = &\sum_i^{N_{pos}}{BCE(p_i, t_i) \times \left[\left(1 - \sqrt{\frac{h_i}{H} \frac{w_i}{W}}\right)^\alpha + 1\right]} \\
    &+ \sum_j^{N_{neg}}{p_j^{2}BCE(p_j, 0)}
    \label{eq:pocoo}
\end{split}
\end{equation}
where $i$ and $j$ index object predictions. $h_i$ and $w_i$ denote the height and width of the matched ground truth box. $p$ and $t$ are the predicted classification score and new label, respectively, and $\alpha$ ranges between 0 and 1.
The key distinction between our PoCoo loss and IA-BCE loss lies in the term $\left[ * \right]$. In Eq~\ref{eq:pocoo}, we incorporate box sizes as the priors into the loss function, explicitly assigning higher weights to small objects and encouraging the model to focus more on them.

%% file: sec/4_experiments.tex
\section{Experiments}
\label{sec:exp_setup}

\quad \textbf{Dataset.} We evaluate DS-Det on two detection benchmark datasets: COCO2017~\cite{lin2014microsoft} and WiderPerson~\cite{zhang2019widerperson}. These two datasets vary in the number of training images and the diversity of detection scenes. Ablation studies are conducted on the COCO2017 dataset.

\textbf{Implementation Details.} For fair comparison, we adopt the same training recipe from DINO~\citep{zhang2022dino}.
DS-Det utilizes \textit{4-scale} features from the backbone. The models are trained with a mini-batch size 8 on Tesla V100 GPUs. In ablations, our models are all trained for 12 epochs (1$\times$ training scheduler).

\textbf{Evaluation Criteria.} For COCO2017, we evaluate performance using the standard average precision (AP) metric under various IoU thresholds and object scales, following evaluation metrics in COCO~\cite{lin2014microsoft}. For WiderPerson, we employ the evaluation metrics of AP, Recall, and mMR, which are commonly used in pedestrian detection~\citep{zhang2019widerperson, rukhovich2021iterdet}.

\subsection{Main Results}
\label{sec:main_results}

\textbf{Results on COCO2017.} Table~\ref{tab:sota} presents a comprehensive comparison of DS-Det with multiple popular detectors using \textit{various backbones} across \textit{different} training epochs. 
DS-Det achieves higher performance across five different backbones~\cite{he2016deep, cao2023strip, liu2021swin, liu2024vmamba} in terms of AP and AP$_S$ for general object detection and small object detection, respectively. With the ResNet50 backbone, DS-Det outperforms the baseline DINO by \textbf{+1.5\%} AP and \textbf{+2.3\%} AP$_S$ under 1 $\times$ scheduler (12 epochs).
Notably, DS-Det (trained for \textbf{24} epochs only) achieves \textbf{+0.4\%} AP (51.3\% vs. 50.9\%) and \textbf{+0.9\%} AP$_S$ (35.5\% vs. 34.6\%) compared to DINO (36 epochs), demonstrating superior training efficiency and effectiveness.
For Strip-MLP-T~\citep{cao2023strip} and Swin-T~\cite{liu2021swin} backbones, DS-Det achieves new results with \textbf{55.0\%} AP and \textbf{54.9\%} AP, respectively.
With the larger Swin-L~\cite{liu2021swin} backbone, DS-Det surpasses DINO by \textbf{+0.9\%} AP (57.7\% vs. 56.8\%). 

Recently, the visual state space model~\cite{gu2023mamba} has been introduced to address the quadratic complexity of the attention mechanism, and MambaOut~\cite{yu2024mambaout} highlights its potential for long-sequence visual tasks like OD. Therefore, we evaluate DS-Det and DINO with the VMamba-T~\cite{liu2024vmamba} backbone. Table~\ref{tab:sota} shows that DS-Det achieves \textbf{54.4\%} AP and \textbf{38.8\%} AP$_S$ under \textbf{1$\times$ }training schedule (12 epochs), surpassing DINO by \textbf{+0.9\%} AP and \textbf{+2.2\%} AP$_S$.
Furthermore, compared to DiffusionDet~\cite{chen2023diffusiondet}, which is capable of predicting a flexible number of objects, DS-Det (24 epochs) significantly outperforms DiffusionDet (60 epochs) with an increase of \textbf{+4.5\%} AP and \textbf{+5.9\%} AP$_S$, demonstrating its superiority and efficiency.

\textbf{Results on WiderPerson.} We conduct experiments on the challenging WiderPerson dataset for further evaluation. Following the recipe of previous works~\cite{zhang2019widerperson, rukhovich2021iterdet}, we report results on the ``Hard'' subset of annotations in Table~\ref{tab:widerperson}. DS-Det outperforms the baseline DINO across all metrics using four different backbones and surpasses other advanced models, reaffirming its effectiveness.

\begin{table}[h]
\small
\centering
\caption{Results on WiderPerson. 
The symbol $^{\dag}$ means the model is trained by us using the official code.}
\resizebox{1.0\columnwidth}{!}{
\setlength{\tabcolsep}{0.5mm}{
 \renewcommand\arraystretch{0.8}
\begin{tabular}{ c c c c c c }
    \toprule
     Method & Year & Epochs & AP$\uparrow$ & Recall$\uparrow$ & mMR$\downarrow$ \\
     \midrule
    PS-RCNN~\citep{ge2020ps} & 2020 & 12 & 89.96 & 94.71 & - \\
    IterDet-2-iter~\citep{rukhovich2021iterdet} & 2021 & 24 & 91.95 & 97.15 & 40.78 \\
    He et al.~\citep{he2022multi} & 2022 & - & 91.29 & - & 40.43 \\
    Cascade Transformer~\citep{ma2023cascade} & 2023 & 50 &  92.98 & 97.66 & 38.41 \\
    \midrule
    DINO-ResNet50$^{\dag}$ & 2022 & 24 & 92.75 & 99.08 & 40.08 \\
    \textbf{DS-Det-ResNet50 (ours)} & 2025 & 24 & \textbf{93.24} & \textbf{99.57} & \textbf{39.47} \\

    \midrule
    DINO-Strip-MLP-T$^{\dag}$ & 2025 & 24 & 93.19 & 99.42 & 38.21 \\
    \textbf{DS-Det-Strip-MLP-T (ours)} & 2025 & 24 & \textbf{93.75} & \textbf{99.65} & \textbf{38.11} \\

    \midrule
    DINO-Swin-T$^{\dag}$ & 2022 & 24 & 93.07 & 99.42 & 38.78 \\
    \textbf{DS-Det-Swin-T (ours)} & 2025 & 24 & \textbf{93.67} & \textbf{99.65} & \textbf{38.05} \\

    \midrule
    DINO-VMamba-T$^{\dag}$ & 2025 & 24 & 93.43 & 99.36 & 38.76 \\
    \textbf{DS-Det-VMamba-T (ours)} & 2025 & 24 & \textbf{94.04} & \textbf{99.65} & \textbf{37.04} \\
     
     \bottomrule
     \end{tabular}}
     }
\label{tab:widerperson}
\end{table}

\subsection{Ablation Studies}

\quad \textbf{Ablations on DS-Det Components.} 
We evaluate the impact of key components in DS-Det using DINO-ResNet50~\cite{zhang2022dino} as the basic model. The baseline model is adapted from fixed-query of DINO to flexible-conditioned query detection by manually setting the query number and removing the SA layers. 
As shown in Table~\ref{tab:ablation_components}, the proposed FLET module, PoCoo loss, and ADD framework effectively enhance performance, demonstrating their effectiveness.
\begin{table}[t]
\centering
\caption{Ablations on DS-Det components. \textit{flexible-c} means the number of queries still constrained by testing parameters.
}
\resizebox{1.0\columnwidth}{!}{
  \renewcommand\arraystretch{0.8}
  \setlength{\tabcolsep}{0.5mm}{
\begin{tabular}{ c c c c  c  c c c c c c c} \\
    \toprule
     FLET & PoCoo & ADD & QNum & AP & AP$_{50}$ & AP$_{75}$ & AP$_S$ & AP$_{M}$ & AP$_{L}$ \\
     \midrule
    ~ & ~ & ~ & fixed & 49.0 & 66.6 & 53.5 & 32.0 & 52.3 & 63.0 \\
    ~ & ~ & ~ & \textit{flexible-c} & 44.1 & 59.5 & 48.1 & 27.9 & 47.6 & 56.7 \\
    \checkmark & ~ & ~ & \textit{flexible} & 48.7 (+4.6) & 65.8 & 53.4 & 31.6 & 52.1 & 62.4 \\
    \checkmark & \checkmark & ~ & \textit{flexible} & 49.3 (+0.6) & 65.6 & 53.9 & 32.2 & 52.6 & 63.4 \\
    \checkmark & \checkmark & \checkmark & \textit{flexible} & \textbf{50.5 (+1.2)} & \textbf{67.5} & \textbf{55.1} & \textbf{34.3} & \textbf{54.6} & \textbf{64.5} \\
     \bottomrule
     \end{tabular}
     }}
\label{tab:ablation_components}
\end{table}

\textbf{Ablations on the Number of SA in DP.} SA plays a critical role in removing duplicate detections. 
We conduct ablations on the number of SA in DP to evaluate their impacts and determine the optimal configuration.
Table~\ref{tab:ablation_sa_num} demonstrates that the absence of SA noticeably decreases performance, dropping to 34.7\% AP, underscoring its necessity in transformer detectors. To balance computational efficiency and accuracy, we utilize 2 layers of SA ($\lambda=2$) in DP for other experiments. Notably, the first four BLP layers for object localization do not include SA, while the last two DP layers each incorporate two SA layers (\textbf{4} in total) for deduplication. This design maintains efficiency, achieving lower computational costs than DINO (which uses \textbf{6} SA modules).
\begin{table}[h]
\centering
\caption{Ablation results on the number of SA layers.}
\resizebox{1.0\columnwidth}{!}{
  \renewcommand\arraystretch{0.8}
  \setlength{\tabcolsep}{4.0mm}{
\begin{tabular}{ c c c c c c c } 
    \toprule
     $\lambda$ & AP & AP$_{50}$ & AP$_{75}$ & AP$_S$ & AP$_{M}$ & AP$_{L}$ \\
     \midrule
     0 & 34.7 & 46.1 & 38.0 & 24.4 & 39.3 & 44.7 \\
     1 & 50.1 & 67.1 & 54.7 & 33.7 & 53.8 & 64.4 \\
     2 & \textbf{50.5} & \textbf{67.5} & \textbf{55.1} & \textbf{34.3} & \textbf{54.6} & 64.5 \\
     3 & 50.4 & 67.5 & 55.1 & 33.6 & 54.1 & \textbf{65.0} \\
     \bottomrule
     \end{tabular}
     }}
\label{tab:ablation_sa_num}
\end{table}

\textbf{Ablations on the PoCoo Loss.} To evaluate the effectiveness of PoCoo loss, we conduct ablations and compare it to the BCE loss and IA-BCE~\cite{cai2023align} loss. The results in Table~\ref{tab:ablation_align_gtszie} indicate that the PoCoo loss outperforms BCE loss and IA-BCE loss by \textbf{+1.7\%} and \textbf{+1.2\%} in AP$_S$, respectively.
In addition, PoCoo loss achieves higher average recall for small objects (AR$_S$), with improvements of \textbf{+0.4\%} and \textbf{+1.0\%} over BCE loss and IA-BCE loss, respectively. These consistent gains in both precision and recall metrics highlight the effectiveness of PoCoo loss for enhancing query learning and performance on hard examples of small objects.
\begin{table}[h]
\centering
\caption{Ablation results on different loss functions.}
\resizebox{1.0\columnwidth}{!}{
  \renewcommand\arraystretch{0.8}
  \setlength{\tabcolsep}{1.0mm}{
\begin{tabular}{ c c c c c c c | c c } 
    \toprule
     Loss Type & AP & AP$_{50}$ & AP$_{75}$ & AP$_S$ & AP$_{M}$ & AP$_{L}$ & AR & AR$_S$ \\
     \midrule
     BCE & 49.0 & 67.3 & 53.4 & 32.6 & 52.4 & 63.0 & 74.3 & 59.6 \\
     IA-BCE & 50.1 & 66.8 & 54.7 & 33.1 & 53.9 & \textbf{65.2} & 74.1 & 59.0 \\
     PoCoo & \textbf{50.5} & \textbf{67.5} & \textbf{55.1} & \textbf{34.3} & \textbf{54.6} & 64.5 & \textbf{74.4} & \textbf{60.0}\\
     \bottomrule
     \end{tabular}
     }}
\label{tab:ablation_align_gtszie}
\end{table}

\textbf{Ablation on PQ.} To further investigate the PQ's impact on DS-Det, we conduct ablations by adding PQ to both the CA and SA modules in ADD. The results presented in Table~\ref{tab:ablation_pos} reveal that our Single-Query paradigm achieves higher performance and that \textit{the PQ is not necessary}. We attribute this to two key factors: (1) Our Single-Query paradigm obtains more comprehensive object information through \textit{encoder-derived tokens} (containing region-specific information where the object itself is located) compared to the solely positional encoding priors of PQ, (2) Through bounding-box-aware point sampling (\textit{e.g.}, deformable attention~\citep{zhu2020deformable}), CA explicitly incorporates location information during query updating, thereby reducing reliance on PQ. Overall, our FLET module streamlines query processing by: simplifying query components, transforming fixed-query into flexible-query, and reducing decoder complexity - enhancing both model adaptability and computational efficiency.
\begin{table}[!h]
\centering
\caption{Ablation results on the PQ in ADD.}
\resizebox{1.0\columnwidth}{!}{
  \renewcommand\arraystretch{1.0}
  \setlength{\tabcolsep}{2.0mm}{
\begin{tabular}{ c c c c c c c c } 
    \toprule
     PQ in SA & PQ in CA & AP & AP$_{50}$ & AP$_{75}$ & AP$_S$ & AP$_{M}$ & AP$_{L}$ \\
     \midrule
     \checkmark & ~ & 50.0 & 67.1 & 54.5 & 33.3 & 53.1 & 65.3 \\
     ~ & \checkmark & 50.3 & 67.4 & 54.8 & 33.5 & 53.7 & 64.5 \\
     \checkmark & \checkmark & 50.1 & 67.1 & 54.5 & 33.5 & 53.9 & \textbf{64.7} \\
     ~ & ~ & \textbf{50.5} & \textbf{67.5} & \textbf{55.1} & \textbf{34.3} & \textbf{54.6} & 64.5 \\
     \bottomrule
     \end{tabular}
     }}
\label{tab:ablation_pos}
\end{table}

\textbf{(6) Ablation on the Stop Gradient of Queries (SGQ).} 
The SGQ plays a crucial role in separating the gradient flow of queries between the one-to-many matching of BLP and the one-to-one matching of DP.
We conduct an ablation on SGQ to show its impact on performance.
The results in Table~\ref{tab:sgq} suggest that the absence of SGQ from DP to BLP leads to a \textbf{2.0\%} decrease in performance, emphasizing the necessity and effectiveness of our SGQ method for decoupled matching to address the issue of query ambiguity.
\begin{table}[h]
\centering
\caption{Ablation results on SGQ.}
\resizebox{1.0\columnwidth}{!}{
  \renewcommand\arraystretch{1.0}
  \setlength{\tabcolsep}{3.0mm}{
\begin{tabular}{ c c c c c c c } 
    \toprule
     SGQ & AP & AP$_{50}$ & AP$_{75}$ & AP$_S$ & AP$_{M}$ & AP$_{L}$ \\
     \midrule
    ~ & 48.5 & 66.5 & 52.9 & 32.2 & 52.1 & 62.5 \\
    \checkmark & \textbf{50.5} & \textbf{67.5} & \textbf{55.1} & \textbf{34.3} & \textbf{54.6} & \textbf{64.5} \\
     \bottomrule
     \end{tabular}
     }}
\label{tab:sgq}
\end{table}

\subsection{Inference Speed Tests and Analysis}
\label{sec:speed}

We conduct additional tests on the inference speed of DS-Det. For a fair comparison, the test was applied on the same codebase (official code of DINO), backbone model (ResNet50), input image size (1280 $\times$ 800), and GPU device (RTX 3090). We test the inference speed of DINO (900 query) and DS-Det (query varies from \textit{10 to 1800}), respectively, as shown in Table~\ref{tab:speed_fps}. Both models share identical backbone and encoder components, resulting in equal processing times. The time of query selection processes is similar ($\sim$7ms for classification scoring). Our FLET module then retrieves queries based on these scores using only \textbf{1 ms}. For 900 queries processing, our ADD decoder achieves \textbf{34.8\%} faster inference than DINO (9.2 ms vs. 14.1 ms), demonstrating its significant efficiency.

\begin{table}[h]
\small
\centering
\caption{Inference speed tests on DINO and DS-Det. }
\resizebox{1.0\columnwidth}{!}{
\setlength{\tabcolsep}{1.0mm}{
 \renewcommand\arraystretch{1.0}
\begin{tabular}{ c | c | c c c c c }
    \toprule
    Model & DINO & DS-Det & DS-Det & DS-Det & DS-Det & DS-Det \\
    Query Number & 900 & 1800 & 900 & 500 & 100 & 10 \\
    \midrule
    Backbone (ms) & 15.2 & 15.2 & 15.2 & 15.2 & 15.2 & 15.2  \\ 
    Encoder (ms) &  30.6 & 30.6 & 30.6 & 30.6 & 30.6 & 30.6 \\
    Query Selection (ms) & 7.3 & 7.8 & 7.5 & 7.4 & 7.4 & 7.4  \\
    \rowcolor{gray!40}{Decoder (ms)} & 14.1 & 11.0 & 9.2 & 8.9 & 8.9 & 8.8 \\
    \midrule
    Inference Time (ms) & 67.2 & 64.6 & 62.5 & 62.1 & 62.1 & 62.0 \\
    FPS (frame/s) & 14.9 & 15.5 & 16.0 & 16.1 & 16.1 & 16.1 \\
     \bottomrule
     \end{tabular}}
     }
\label{tab:speed_fps}
\end{table}

%% file: sec/5_conclusion.tex
\section{Conclusion}
\label{sec:conclusion}

This paper proposes DS-Det, a novel efficient detector capable of detecting a flexible number of objects in different images. DS-Det reformulates the query paradigm by introducing a Single-Query type with FLET module, transforming the fixed-query into flexible-query. By rethinking the roles of attention mechanisms, we disentangle the cross-attention and self-attention layers and further design a new efficient ADD architecture, combined with decoupled one-to-many and one-to-one matching approaches, addressing the ``query ambiguit'' and ``ROT'' issues. Extensive experiments on diverse datasets demonstrate the effectiveness of DS-Det across various backbones and significantly improve decoder inference speed. We hope that DS-Det inspires the development of high-quality object detectors and multi-modal models in future research.

%% file: sec/appendix.tex
\onecolumn
\appendix

\section{Appendix for Methodology}
\label{sec:app_methods}

\subsection{Main Pipeline of DETR-like Models.}
\label{sec:app_main_pipeline_detr}

The DETR-like architecture operates as an end-to-end framework, where each ground truth object is assigned to a single positive query, following a one-to-one manner. Specifically, this architecture comprises three main modules: a compact backbone for feature extraction, a transformer encoder neck for feature enhancement, and a transformer-decoder for predicting bounding boxes and classes. 
Given an input image $I \in \mathbb{R}^{H \times W \times 3}$ (H, W: image height and width), the backbone extracts a compact feature representation $\bm{B}$. This feature is then passed through the transformer encoder, which consists of a chain of attention~\cite{dosovitskiy2020image} or deformable attention~\cite{zhu2020deformable} layers to disentangle objects and obtain the encoder feature $\bm{E}$. Next, the model initializes a fixed-size set of two types of queries: content query (CQ) and positional query (PQ)~\cite{liu2022dab, zhang2022dino}. These queries, along with the encoder feature $\bm{E}$, are fed into the transformer decoder, which further updates these queries based on the information from the encoder feature via the SA and CA modules. Finally, the content queries are passed through two separate \textbf{f}eed \textbf{f}orward \textbf{n}etworks (FFN) to predict the bounding box coordinates and class labels, respectively. 

\subsection{How Does the SA Module Reduce Duplicate Detections?}
\label{app:sa_supervision}

In object detection, it is extremely challenging~\cite{carion2020end, cheng2023towards} to deduplicate detected objects. Non-maximum suppression (NMS) is conventionally employed as a post-processing to remove duplicate bounding boxes based on overlap (IoU). However, this approach relies on manual thresholds and may mistakenly remove overlapping objects, hurting performance. In transformer detectors, where each query predicts only one object, \textit{the similarity between queries can be used to deduplicate the predictions}, enabling end-to-end training.
This is done by using query similarities as attention scores in SA, derived from \textit{pairwise} attention maps between all queries.
For these similar queries, with the one-to-one matching label alignment mechanism, each query matched to a ground truth box will gradually obtain a higher classification score under the supervision of the classification loss function. In contrast, unmatched queries will get progressively lower scores. 
Ultimately, by reducing the classification scores of similar queries, SA can effectively deduplicate the predictions.

\subsection{Complexity Analysis}
\label{sec:complexity_analysis_1}

DS-Det is a novel flexible detector that can adaptively select a variable number of queries for different input images, as shown in Fig.~\ref{fig:vis1}, Fig.~\ref{fig:vis2}, and Fig.~\ref{fig:vis3}. 
Due to the dynamic computational complexity resulting from the flexible query selection process, we use a fixed 900 queries for DS-Det to calculate the FLOPs in Table ~\textcolor{purple}{3} of the main text for fair comparison with other models.

\begin{figure}[ht]
  \centering
   \includegraphics[width=0.5\linewidth]{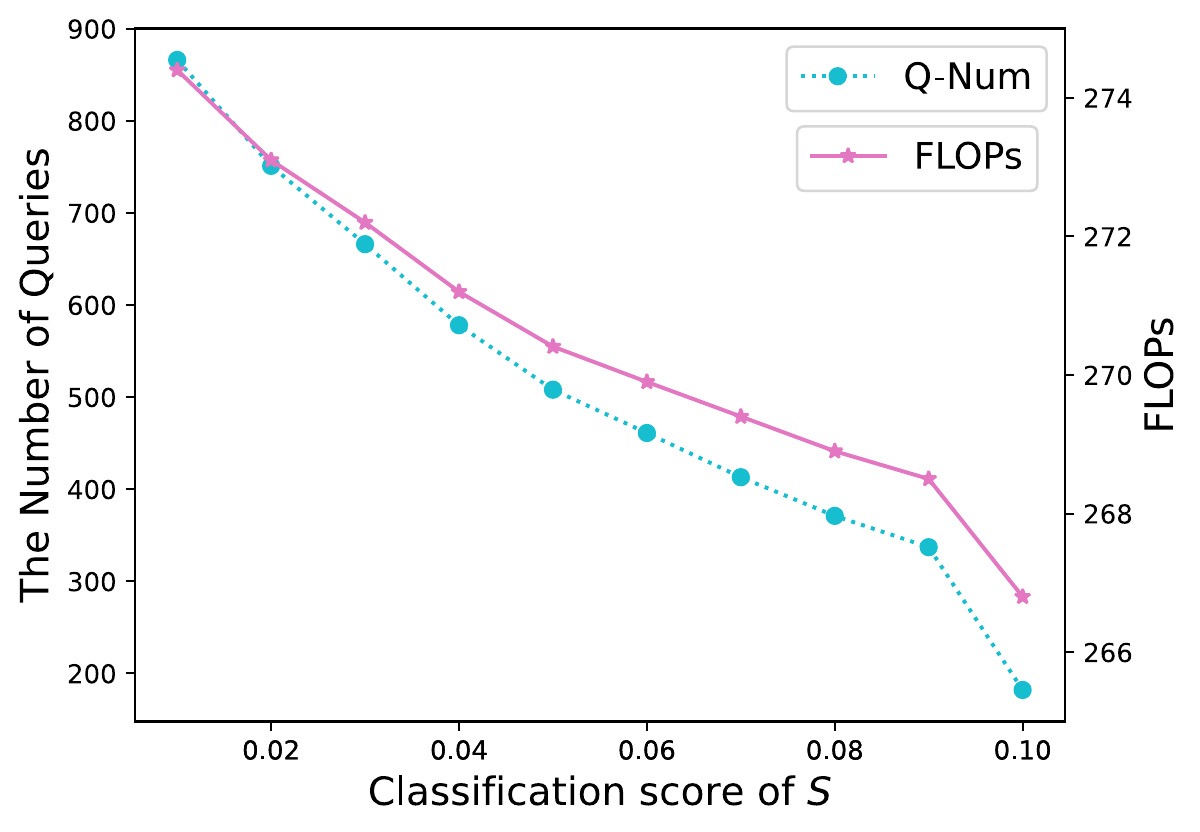}
   \caption{The number of queries and FLOPs of the model w.r.t threshold of classification score $S$.}
   \label{fig:flops}
\end{figure}

Actually, the classification threshold $S$ affects the number of selected queries: a higher $S$ leads to fewer queries and vice versa. Taking the DS-Det-ResNet50 model with a pool size of 900 as an example, our model can effectively reduce the number of queries, and the corresponding FLOPs are also reduced, as shown in Fig.~\ref{fig:flops}. The ablation results on the $S$ are presented in Table~\ref{tab:ablation_threshold} of Sec.~\ref{app:ablation}.

\section{Appendix for Experiments}
\label{app:ablations_exp}

\subsection{Details about Datasets}
\label{app:dataset}

\quad \textbf{COCO2017.} The COCO2017~\citep{lin2014microsoft} dataset is a widely used benchmark dataset for object detection. 
It consists of 118k training images and 5k validation images, with 80 object categories.

\textbf{WiderPerson.} WiderPerson~\citep{zhang2019widerperson} is a large and diverse dataset for dense pedestrian detection in real-world settings. It consists of 13,382 images with a total number of 399,786 annotations, averaging 29.87 annotations per image. This dataset presents significant challenges for SOD due to its diverse scenarios and substantial occlusion. It includes 8,000 images for training and 1,000 images for validation.

\subsection{Experimental Tests}
\label{sec:experimental_test}

\begin{figure}[h]
\centering
\includegraphics[width=0.5\textwidth]{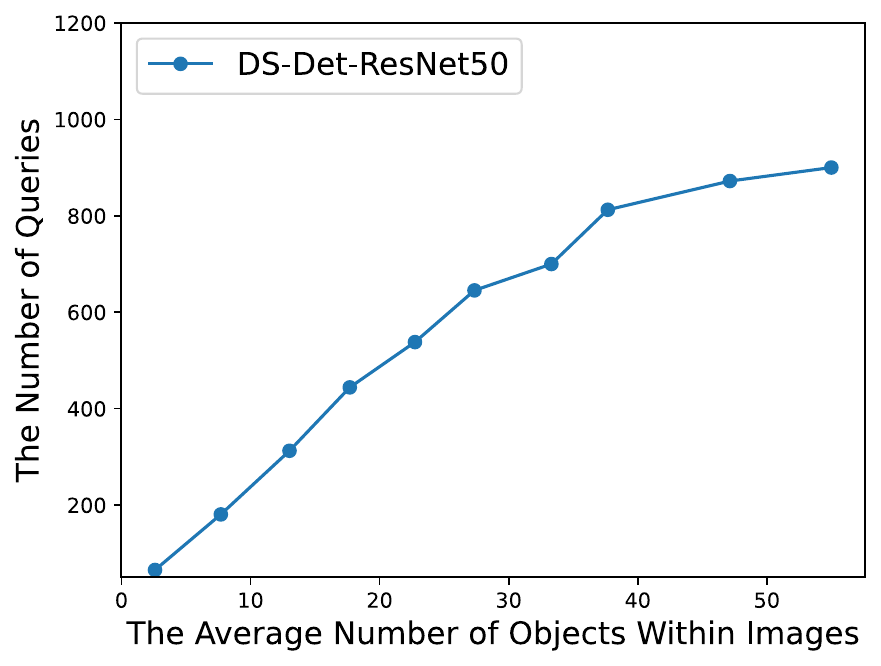} 
\caption{The relation between the number of objects and queries dynamically generated by the DS-Det.
}
\label{fig:query_trend}
\end{figure}

\begin{figure}[h]
\centering
\includegraphics[width=0.43\textwidth]{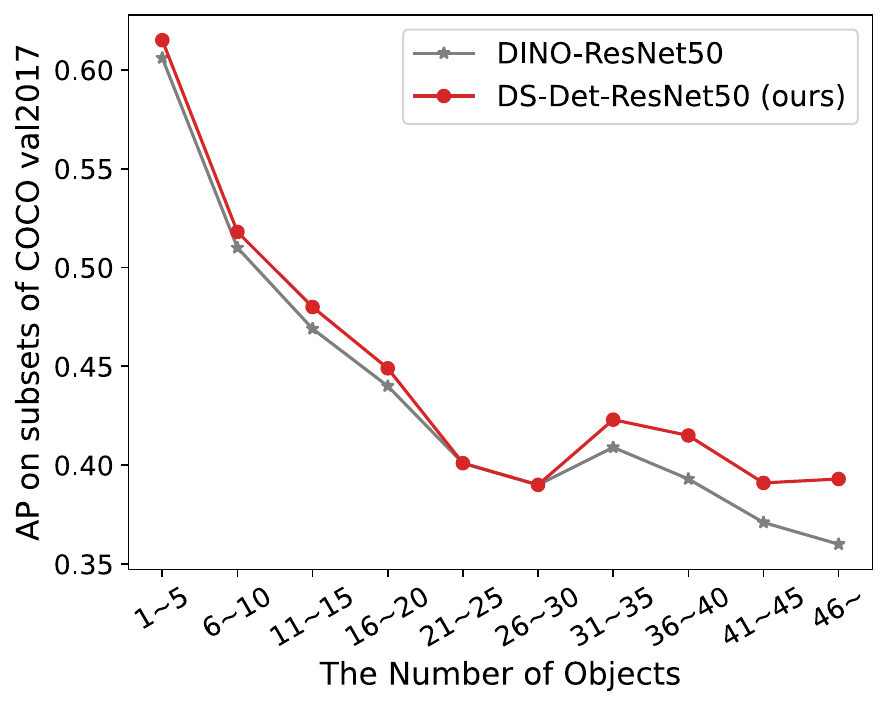} 
\caption{The performance comparison on the subsets of val2017 of COCO. The subsets are generated based on the number of objects per image with a step size of 5.
}
\label{fig:query_acc}
\end{figure}

\textbf{The proportional relation between the number of objects and queries.} It is intuitive that the more potential objects to be detected, the more queries would be required. To verify the effectiveness of FLET in flexible query generation, we conducted an additional test to observe the trends in model accuracy and the dynamic generation of query quantity with the number of objects in the test images increases. Specifically, we divided the COCO validation set into 10 subsets based on the number of objects in each image, with a step size of 5 objects per image. Then, we tested the performance and counted the number of queries generated by the DS-Det model for each subset. As shown in Table ~\ref{tab:relation}, the number of queries increases as the number of objects to be detected increases, and the growth rate gradually becomes slow, as illustrated in Fig.~\ref{fig:query_trend}. At the same time, our model obtained overall higher performance across all subsets, further validating its effectiveness, as presented in Table ~\ref{tab:relation} and Fig.~\ref{fig:query_acc}.

When processing images with more objects, the advantages of our method become more apparent, outperforming DINO by +2.0\% AP and +3.3\% AP in the subsets of 41-45 and over 46, respectively. The trend of the query number starts to slow down as the number of objects in the image increases, as shown in Fig.~\ref{fig:query_trend}.
The subset of 1-5 occupies 55.9\% images among val2017 of COCO. For this subset, our model only uses \textbf{7.25\%} of the queries (65.22 vs 900) while achieving higher performance by \textbf{+0.9\% AP}, indicating that the selected queries via FLET module are more effective. This demonstrates a significant advantage in common scenarios, as it can effectively reduce computational costs. Moreover, the number of queries selected by our model can be adaptively adjusted based on different classification thresholds, without the need to retrain the model, making it more easily adaptable to different detection scenarios.

\begin{table*}[h]
\small
\centering
\caption{The performance (AP) on the subsets of val2017 of COCO.
}
\resizebox{1.0\columnwidth}{!}{
\setlength{\tabcolsep}{2.0mm}{
 \renewcommand\arraystretch{1.0}
\begin{tabular}{ c | c c c c c c c c c c c }
    \toprule
    Object Numbers & 1-5 & 6-10 & 11-15 & 16-20 & 21-25 & 26-30 & 31-35 & 36-40 & 41-45 & 46+ \\
    \midrule
     Average Objects & 2.60 & 7.70 & 13.02 & 17.69 & 22.74 & 27.35 & 33.31 & 37.68 & 42.77 & 55.00 \\
     Image Numbers & 2769 & 985 & 556 & 324 & 170 & 83 & 29 & 22 & 9 & 5 \\
    Query Numbers & 65.22 & 180.46 & 312.47 & 443.98 & 537.91 & 645.16 & 699.82 & 812.41 & 872.00 & 900.00 \\
    Query Num/Object Num & 25.08 & 23.44 & 23.99 & 25.10 & 23.65 & 23.59 & 21.01 & 21.56 & 20.39 & 16.37 \\
     \midrule
    DINO & 60.6 & 51.0 & 46.9 & 44.0 & 40.1 & 39.0 & 40.9 & 39.3 & 37.1 & 36.0 \\
    \textbf{DS-Det (ours)} & \textbf{61.5} & \textbf{51.8} & \textbf{48.0} & \textbf{44.9} & \textbf{40.1} & \textbf{39.0} & \textbf{42.3} & \textbf{41.5} & \textbf{39.1} & \textbf{39.3} \\
     \bottomrule
     \end{tabular}}
     }
\label{tab:relation}
\end{table*}

\begin{figure}[h]
\centering
\includegraphics[width=0.46\textwidth]{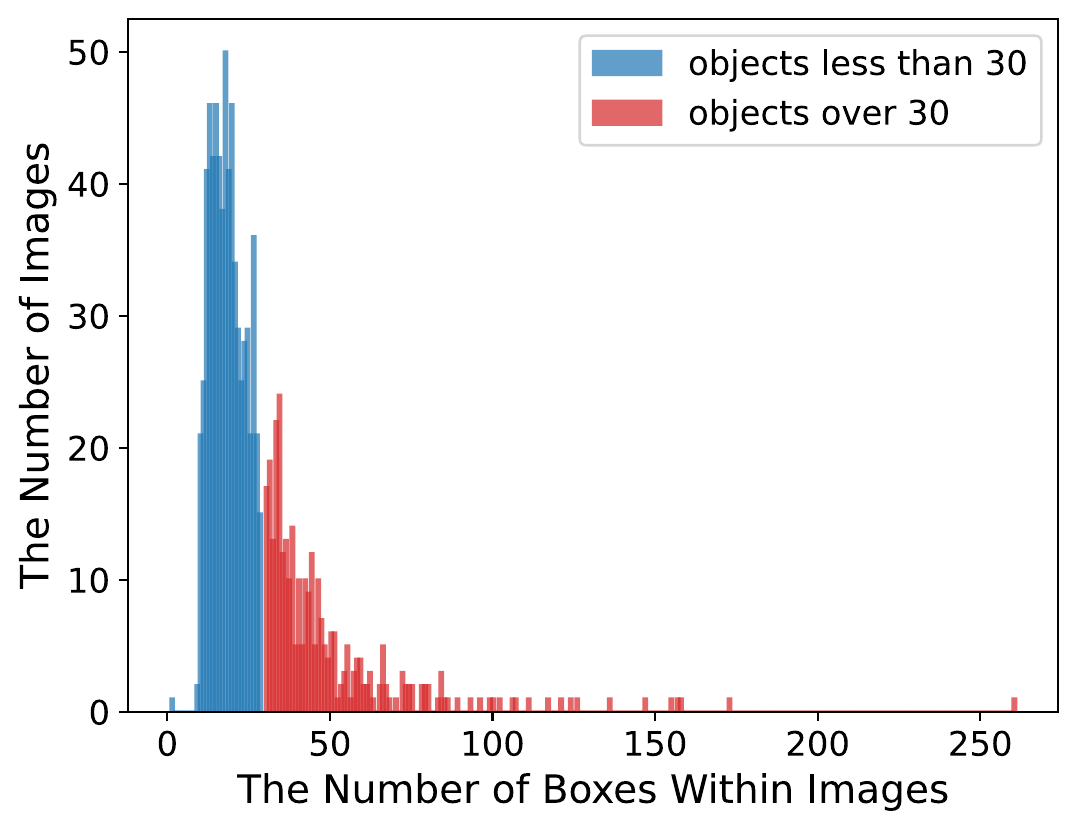} 
\caption{The histogram of the WiderPerson validation dataset.
}
\label{fig:wid_sts}
\end{figure}

\textbf{Performance on challenging dense objects detection of WiderPerson dataset.} WiderPerson is a large, diverse, and challenging dataset for dense pedestrian detection, with an average of 29.87 annotations per image. As illustrated in Fig.~\ref{fig:wid_sts}, the statistic results on its validation set indicate that there are 679 images with less than 30 objects, and 321 images with 30 or more objects. To evaluate the effectiveness of our method in handling the more challenging scenario with dense objects over 30 of the WiderPerson dataset, we divided this validation set into two test subsets. The results in Table~\ref{tab:widerperson_30} demonstrate that, across four backbone models, our models consistently achieve overall higher AP, Recall, and lower mMR, further validating its effectiveness.

\begin{table*}[!h]
\small
\centering
\caption{The experimental results on the WiderPerson dataset.}
\resizebox{1.0\columnwidth}{!}{
\setlength{\tabcolsep}{5.0mm}{
 \renewcommand\arraystretch{1.0}
\begin{tabular}{ c c c c | c c c c}
    \toprule
    Objects Per Image & \multicolumn{3}{c|}{Less Than 30 (679 images)} & \multicolumn{3}{c}{Over 30} (321 images)\\
    
    \midrule    
     Method & AP$\uparrow$ & Recall$\uparrow$ & mMR$\downarrow$ & AP$\uparrow$ & Recall$\uparrow$ & mMR$\downarrow$ \\
     
     \midrule

    DINO-ResNet50 & 95.84 & 99.63 & \textbf{26.71} & 88.84 & 98.45 & 57.18 \\
    \textbf{DS-Det-ResNet50 (ours)} & \textbf{96.10} & \textbf{99.80} & 27.50 & \textbf{89.65} &\textbf{99.31} & \textbf{55.09} \\

    \midrule
    DINO-Strip-MLP-T & 96.27 & 99.80 & 26.16 & 90.07 & 99.40 & 56.33 \\
    \textbf{DS-Det-Strip-MLP-T (ours)} & \textbf{96.39} & \textbf{99.83} & \textbf{26.03} & \textbf{90.45} & \textbf{99.45} & \textbf{53.37} \\

    \midrule
    DINO-Swin-T & 96.45 & 99.71 & \textbf{26.35} & 90.06 & 99.00 & 55.58 \\
    \textbf{DS-Det-Swin-T (ours)} & \textbf{96.52} & \textbf{99.84} & 26.36 & \textbf{90.08} & \textbf{99.43} & \textbf{53.63} \\

    \midrule
    DINO-VMamba-T & 96.09 & \textbf{99.83} & 27.12 & 90.09 & 99.46 & 53.68 \\
    \textbf{DS-Det-VMamba-T (ours)} & \textbf{96.60} & 99.80 & \textbf{25.68} & \textbf{90.76} & \textbf{99.47} & \textbf{52.08}\\
     
     \bottomrule
     \end{tabular}}
     }
\label{tab:widerperson_30}
\end{table*}

\subsection{Experiment on Other Transformer Detector} 

\begin{figure}[h]
\centering
\includegraphics[width=0.5\textwidth]{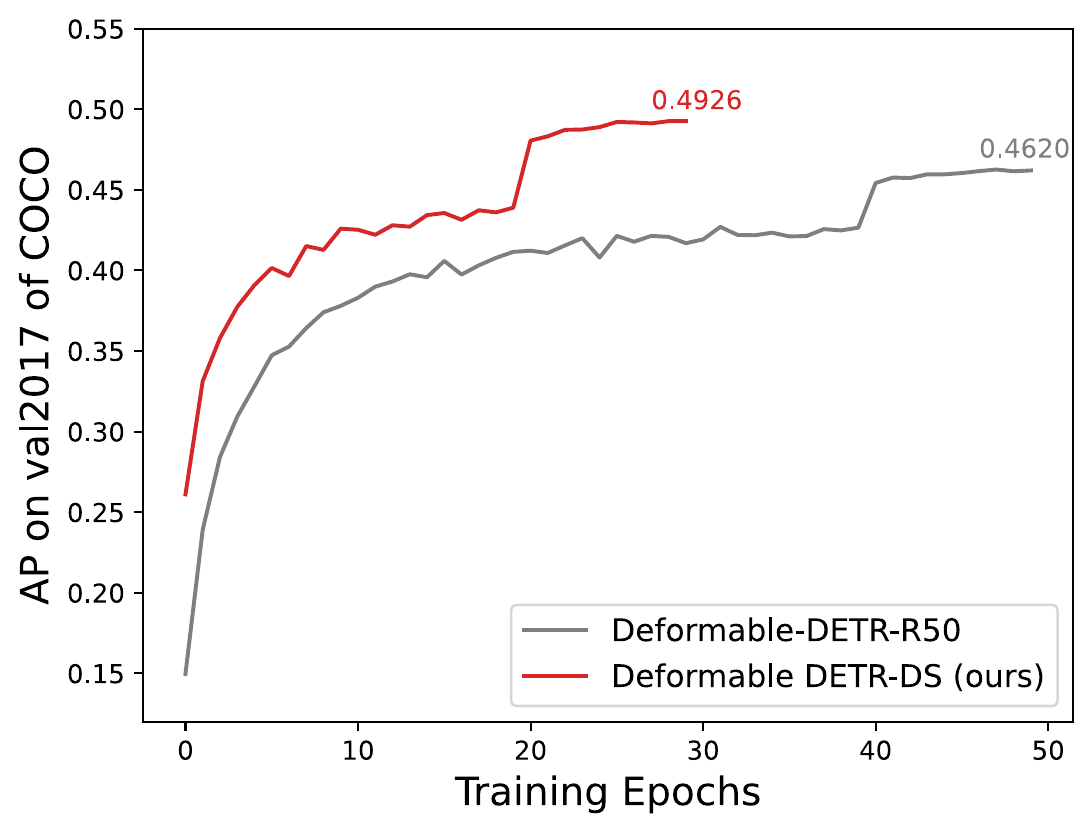} 
\caption{Convergence curves of Deformable DETR and Deformable DETR-DS model. 
}
\label{fig:deformable_train}
\end{figure}

To demonstrate the generalizability of DS-Det beyond DINO, we conduct experiments with another widely used detector, Deformable DETR~\cite{zhu2020deformable}. Specifically, we introduce a new variant, Deformable DETR-DS, by integrating FLET, PoCoo loss, and the ADD architecture into Deformable DETR. The results in Table ~\ref{tab:deformable} show that our model outperforms the original Deformable DETR by \textbf{+3.1\%} AP and \textbf{+3.5\%} AP$_S$ on the COCO dataset and converges faster (in Fig.~\ref{fig:deformable_train}), further validating its effectiveness and generalizability.

\begin{table}[h]
\centering
\caption{Results of the Deformable DETR and its variants with our approaches on val2017 of COCO dataset.}
\resizebox{1.0\columnwidth}{!}{
  \renewcommand\arraystretch{1.0}
  \setlength{\tabcolsep}{4.0mm}{
\begin{tabular}{ c | c | c c c c c c}
    \toprule
     Method & Eps & AP &  AP$_{50}$ & AP$_{75}$ & AP$_S$ & AP$_{M}$ & AP$_{L}$ \\
     \midrule
     Deformable DETR~\cite{zhu2020deformable} & 50 & 46.2 & 65.2 & 50.0 & 28.8 & 49.2 & 61.7 \\
    
     \textbf{Deformable DETR-DS (ours)} & \textbf{30} & \textbf{49.3 (+3.1)} & \textbf{66.4} & \textbf{53.6} & \textbf{32.3 (+3.5)} & \textbf{52.7} & \textbf{63.9} \\
     
     \bottomrule
     \end{tabular}
     }}
\label{tab:deformable}
\end{table}

\subsection{Ablation Studies}
\label{app:ablation}


\quad \textbf{(1) Ablation on the classification threshold $S$ for FLET module.} The classification threshold $S$ in FLET module enables adaptive control of the number of decoder queries. A higher value of $S$ reduces the number of decoder queries, while a lower $S$ increases them, as illustrated in Fig.~\ref{fig:flops}. 
We conduct ablation experiments with varying $S$ to evaluate its impact on the model performance, as shown in Table~\ref{tab:ablation_threshold}.
The smaller $S$ brings higher performance, and when $S$ ranges from $0.01$ to $0.05$, the model performance varies in a small range of $50.2\% \sim 50.6\%$ AP, indicating good robustness.
Notably, when $S=0.10$, our model used only 20.2\% queries of the DINO on average, yet it achieved a \textbf{+0.5\%} higher AP than DINO (49. 5\% vs 49. 0\%).
These results demonstrate the significant effectiveness of our FLET module for ``flexible object predictions''.


\begin{table*}[h]
\centering
\caption{Ablation results of threshold of $S$ in FLET module.}
\resizebox{1.0\columnwidth}{!}{
\setlength{\tabcolsep}{10.0mm}{
 \renewcommand\arraystretch{1.0}
\begin{tabular}{ c c c c c c c}
    \toprule
     $S$ & AP & AP$_{50}$ & AP$_{75}$ & AP$_S$ & AP$_{M}$ & AP$_{L}$ \\
     \midrule
     0.01 & \textbf{50.6} & \textbf{67.9} & \textbf{55.4} & 34.1 & 54.3 & 64.8 \\
     0.02 & 50.5 & 67.5 & 55.1 & \textbf{34.3} & \textbf{54.6} & 64.5 \\
     0.03 & 50.2 & 67.4 & 54.6 & 33.2 & 53.7 & 64.8 \\
     0.04 & 50.2 & 67.2 & 54.5 & 33.6 & 53.8 & 64.5 \\
     0.05 & 50.3 & 67.5 & 54.8 & 33.4 & 54.0 & \textbf{65.0} \\
     \midrule
     0.06 & 50.1 & 67.3 & 54.6 & 33.2 & 53.7 & 64.5 \\
     0.07 & 49.8 & 66.7 & 54.5 & 32.8 & 53.4 & 64.1 \\
     0.08 & 50.0 & 67.0 & 54.5 & 33.2 & 53.9 & 64.0 \\
     0.09 & 49.6 & 66.7 & 54.0 & 32.1 & 53.3 & 63.8 \\
     0.10 & 49.5 & 66.4 & 54.0 & 33.2 & 53.0 & 64.0 \\
     \bottomrule
     \end{tabular}
     }}
\label{tab:ablation_threshold}
\end{table*}

\textbf{(2) The architecture configuration of ADD.} To maintain consistency with methods such as DETR and DINO, we design ADD architecture with six decoder layers, ensuring that the parameter count remains consistent. However, the distribution of BLP and DP within these six layers affects the model's performance: excessive BLP layers can hinder the model's ability to eliminate duplicate detections effectively; conversely, an excessive number of DP layers may lead to inaccurate bounding box predictions and decreased training efficiency. To determine the optimal configuration, we perform ablations on the number of BLP and DP layers in ADD. As presented in Table~\ref{tab:ablation_blp_dp}, the best performance across all metrics is achieved with 4 BLP and 2 DP layers.
We adopt this configuration for other experiments.

\begin{table}[h]
\centering
\caption{Ablation on the BLP and DP layers.}
\resizebox{1.0\columnwidth}{!}{
  \renewcommand\arraystretch{1.0}
  \setlength{\tabcolsep}{8.0mm}{
\begin{tabular}{ c c c c c c c c } 
    \toprule
     BLP & DP & AP & AP$_{50}$ & AP$_{75}$ & AP$_S$ & AP$_{M}$ & AP$_{L}$ \\
     \midrule
     1 & 5 & 49.5 & 67.1 & 54.2 & 33.3 & 53.1 & 63.5 \\
     2 & 4 & 49.4 & 67.2 & 53.8 & 33.4 & 52.9 & 63.3 \\
     3 & 3 & 50.0 & 67.1 & 54.6 & 33.3 & 53.7 & 64.5 \\
     \textbf{4} & \textbf{2} & \textbf{50.5} & \textbf{67.5} & \textbf{55.1} & \textbf{34.3} & \textbf{54.6} & 64.5 \\
     5 & 1 & 50.0 & 66.7 & 54.3 & 32.9 & 53.8 & \textbf{64.7} \\
     \bottomrule
     \end{tabular}
     }}
\label{tab:ablation_blp_dp}
\end{table}


\textbf{(3) Ablation on the connection order of CA and SA in DP.} 
In existing transformer-based detectors, it is commonly observed that SA is connected before CA in the decoder, following the original architecture in DETR.
However, this study argues that this connection scheme may introduce query ambiguity due to the opposing impacts of SA and CA on the object queries. We conduct an ablation study on DS-Det to investigate the effect of reversing the order of CA and SA connections. Table~\ref{tab:ablation_reverse} demonstrates the effectiveness of our connection scheme of SA in DP, significantly enhancing the model's ability to remove duplicate detections.

\begin{table}[!h]
\centering
\caption{Ablation on the order of the SA and CA.}
\resizebox{1.0\columnwidth}{!}{
  \renewcommand\arraystretch{1.0}
  \setlength{\tabcolsep}{8.0mm}{
\begin{tabular}{ c c c c c c c } 
    \toprule
     Connection & AP & AP$_{50}$ & AP$_{75}$ & AP$_S$ & AP$_{M}$ & AP$_{L}$ \\
     \midrule
     SA $\rightarrow$ CA & 50.1 & 67.1 & 54.6 & 33.5 & 53.3 & \textbf{64.7} \\
     \textbf{CA $\rightarrow$ SA} & \textbf{50.5} & \textbf{67.5} & \textbf{55.1} & \textbf{34.3} & \textbf{54.6} & 64.5 \\
     \bottomrule
     \end{tabular}
     }}
\label{tab:ablation_reverse}
\end{table} 

\textbf{(4) Ablation on the one-to-many matching of $K$.} One-to-many matching is a significant approach to enhance training efficiency by increasing the number of positive samples. We perform ablations on the ground truth box, repeating times $K$ to determine the optimal configuration.
As $K$ increases, removing duplicate box detections becomes more difficult. Conversely, smaller $K$ results in insufficient positive samples, decreasing training efficiency.
Table~\ref{tab:ablation_one_to_many_k} shows the best performance is obtained with $K=6$, which is used for other experiments.

\begin{table}[h]
\centering
\caption{Ablation on K of one-to-many matching.}
\resizebox{1.0\columnwidth}{!}{
  \renewcommand\arraystretch{1.0}
  \setlength{\tabcolsep}{8.0mm}{
\begin{tabular}{ c c c c c c c } 
    \toprule
     $K$ & AP & AP$_{50}$ & AP$_{75}$ & AP$_S$ & AP$_{M}$ & AP$_{L}$ \\
     \midrule
     1 & 50.1 & 67.0 & 54.6 & 33.4 & 53.7 & 64.9 \\
     3 & 50.2 & 67.2 & 54.5 & 33.4 & 54.1 & 64.7 \\
     \textbf{6} & \textbf{50.5} & \textbf{67.5} & \textbf{55.1} & \textbf{34.3} & \textbf{54.6} & 64.5 \\
     9 & 50.1 & 66.8 & 54.7 & 33.1 & 53.9 & \textbf{65.2} \\
    12 & 49.8 & 67.0 & 54.1 & 32.6 & 53.6 & 64.4 \\
     \bottomrule
     \end{tabular}
     }}
\label{tab:ablation_one_to_many_k}
\end{table}

\textbf{(5) Ablation on the classification cost weight for the matching process.} Accurately matching predicted boxes with ground truth boxes is critical for transformer-based models. We employ the same cost components as DETR and DINO, including the L1 cost for bounding boxes, binary cross-entropy (BCE) cost for classification, and generalized Intersection over Union (GIoU) cost. 
The weights assigned to each cost component play a significant role in optimizing the training process and affecting the model's performance.
In our DS-Det model, decoupled matching primarily addresses the issue of duplicate detections through the one-to-one matching classification supervision and the DP module. 
To investigate the impact of different \textit{classification costs} on model training, we conduct ablation experiments to determine the optimal configurations. The results in Table~\ref{tab:ablation_cost_weight} show that the absence of the classification cost (BCE$_{BLP}=0.0$) hinders the model's performance. 
\textit{Including the classification cost in the matching process can introduce semantic information}, leading to more appropriate query matches.
However, assigning a higher weight to the classification cost makes it more challenging for the model to predict bounding boxes accurately, as the query with a higher classification score would be matched with the ground truth box rather than the query with a higher IoU.
Based on the experimental results, we adopt a weight of 0.2 for BCE cost as our training parameter.

\begin{table*}[h]
\centering
\caption{Ablation on the classification cost weight in BLP.}
\resizebox{1.0\columnwidth}{!}{
  \renewcommand\arraystretch{1.0}
  \setlength{\tabcolsep}{2.0mm}{
\begin{tabular}{ c  c c c c c  c c c c c c } 
    \toprule
      BCE$_{BLP}$ & L1$_{BLP}$ & GIou$_{BLP}$ & BCE$_{DP}$ & L1$_{DP}$ & GIou$_{DP}$ & AP & AP$_{50}$ & AP$_{75}$ & AP$_S$ & AP$_{M}$ & AP$_{L}$ \\
     \midrule
     0.0 & 5.0 & 2.0 & 2.0 & 2.0 & 2.0 & 48.9 & 66.0 & 52.9 & 31.3 & 52.6 & 63.3 \\
     0.1 & 5.0 & 2.0 & 2.0 & 2.0 & 2.0 & 50.0 & 67.0 & 54.6 & 33.4 & 53.4 & 64.8 \\
     \textbf{0.2} & 5.0 & 2.0 & 2.0 & 2.0 & 2.0 & \textbf{50.5} & 67.5 & \textbf{55.1} & \textbf{34.3} & \textbf{54.6} & 64.5 \\
     0.3 & 5.0 & 2.0 & 2.0 & 2.0 & 2.0 & 49.9 & 67.0 & 54.3 & 33.3 & 53.5 & 64.4 \\
     0.5 & 5.0 & 2.0 & 2.0 & 2.0 & 2.0 & 50.2 & \textbf{67.7} & 54.7 & 34.0 & 53.7 & \textbf{64.8} \\
     0.8 & 5.0 & 2.0 & 2.0 & 2.0 & 2.0 & 49.4 & 66.9 & 53.7 & 33.1 & 52.9 & 64.4 \\
     \bottomrule
     \end{tabular}
     }}
\label{tab:ablation_cost_weight}
\end{table*}

\textbf{(6) Ablation on the loss weight for BLP and DP.} In transformer-based detectors~\cite{carion2020end}, there are three main loss functions: BCE loss for classification, L1 and GIoU loss for bounding box regression.
To investigate the impact of different loss weights on the encoder, BLP, and DP components of the model, we conduct ablation experiments using various weight values.
As presented in Table~\ref{tab:ablation_loss_weight}, the baseline model (index 0) is achieved with the same weight as DINO. 
To improve the one-to-one classification accuracy of the model, we reduce the weight of the L1 loss in the DP component (index 1) and increase the weight of the classification loss (index 2).
Building upon the baseline, we further increase the overall weight of the classification loss across the encoder, BLP, and DP (index 3). 
We then test the effect of increasing the weights of both the classification loss and the GIoU loss (index 4, 5, and 6, respectively). Experiments of index 3 and 6 indicate that a higher weight on the L1 loss of bounding boxes is important for box refinement in the DP component.
Finally, we adopt the weight configuration of index 3 as the loss weights for other model training.

\begin{table*}[h]
\centering
\caption{Ablation on different loss weights for encoder, BLP, and DP.}
\resizebox{1.0\columnwidth}{!}{
  \renewcommand\arraystretch{1.0}
  \setlength{\tabcolsep}{1.0mm}{
\begin{tabular}{ c c c c c c c c  c c c c c c } 
    \toprule
      Index & PoCoo$_{enc}$ & PoCoo$_{BLP}$ & L1$_{BLP}$ & GIou$_{BLP}$ & PoCoo$_{DP}$ & L1$_{DP}$ & GIou$_{DP}$ & AP & AP$_{50}$ & AP$_{75}$ & AP$_S$ & AP$_{M}$ & AP$_{L}$ \\
     \midrule
     0 (baseline) & 1.0 & 1.0 & 5.0 & 2.0 & 1.0 & 5.0 & 2.0 & 49.4 & 66.2 & 54.1 & 32.4 & 53.1 & 64.2 \\
     1 & 1.0 & 1.0 & 5.0 & 2.0 & 1.0 & 1.0 & 2.0 & 49.5 & 66.4 & 54.0 & 32.3 & 53.4 & 63.5 \\
     2 & 1.0 & 1.0 & 5.0 & 2.0 & 2.0 & 1.0 & 2.0 & 48.1 & 64.8 & 52.3 & 31.5 & 52.2 & 62.1 \\
     \midrule
     
     \textbf{3} & 1.5 & 2.0 & 5.0 & 2.0 & 2.0 & 5.0 & 2.0 & \textbf{50.5} & \textbf{67.5} & \textbf{55.1} & \textbf{34.3} & \textbf{54.6} & 64.5 \\
     4 & 1.5 & 3.0 & 5.0 & 2.0 & 3.0 & 1.0 & 2.0 & 49.8 & 67.1 & 54.2 & 33.1 & 53.6 & 64.2 \\
     5 & 1.5 & 3.0 & 5.0 & 3.0 & 3.0 & 2.0 & 3.0 & 49.9 & 67.0 & 54.2 & 32.8 & 53.8 & 64.4 \\
     6 & 1.5 & 2.0 & 5.0 & 3.0 & 2.0 & 5.0 & 3.0 & 50.4 & 67.4 & 55.0 & 33.7 & 54.1 & \textbf{64.6} \\
     \bottomrule
     \end{tabular}
     }}
\label{tab:ablation_loss_weight}
\end{table*}

\textbf{(7) Experiments on the CrowdHuman dataset.} To further demonstrate the effectiveness of our model, we conducted a new experiment on the CrowdHuman~\cite{shao2018crowdhuman} dataset, which is also a challenging dataset for dense pedestrian detection in the wild. The results listed in Table~\ref{tab:crowdhuman} show that DS-Det obtains overall higher performance compared to DINO variants, further confirming the effectiveness of our approach.

\begin{table}[h]
\small
\centering
\caption{The experiment results on the CrowdHuman dataset with full-body annotations.}
\resizebox{1.0\columnwidth}{!}{
\setlength{\tabcolsep}{8.5mm}{
 \renewcommand\arraystretch{1.0}
\begin{tabular}{ c c c c c }
    \toprule
     Method & Epochs & AP$\uparrow$ & Recall$\uparrow$ & mMR$\downarrow$ \\
    \midrule
    DINO-ResNet50 & 24 & 86.61 & 95.11 & 52.85 \\
    \textbf{DS-Det-ResNet50 (ours)} & 24 & \textbf{86.87} & \textbf{95.25} & \textbf{52.08} \\

    \midrule
    DINO-Strip-MLP-T & 24 & \textbf{88.38} & 95.82 & 50.30 \\
    \textbf{DS-Det-Strip-MLP-T (ours)} & 24 & 87.92 & \textbf{95.99} & \textbf{49.83} \\

    \midrule
    DINO-Swin-T & 24 & 87.71 & 95.71 & 51.81 \\
    \textbf{DS-Det-Swin-T (ours)} & 24 & \textbf{88.14} & \textbf{95.86} & \textbf{51.12} \\

    \midrule
    DINO-VMamba-T & 24 & 87.44 & 95.45 & 51.54 \\
    \textbf{DS-Det-VMamba-T (ours)} & 24 & \textbf{88.46} & \textbf{96.20} & \textbf{50.28} \\
     
     \bottomrule
     \end{tabular}}
     }
\label{tab:crowdhuman}
\end{table}

\subsection{Performance Comparison on COCO}

\begin{figure*}[ht]
\small
 \centering
  {
		\begin{minipage}[t]{0.24\linewidth}
			\centering
			\includegraphics[width=0.99\linewidth]{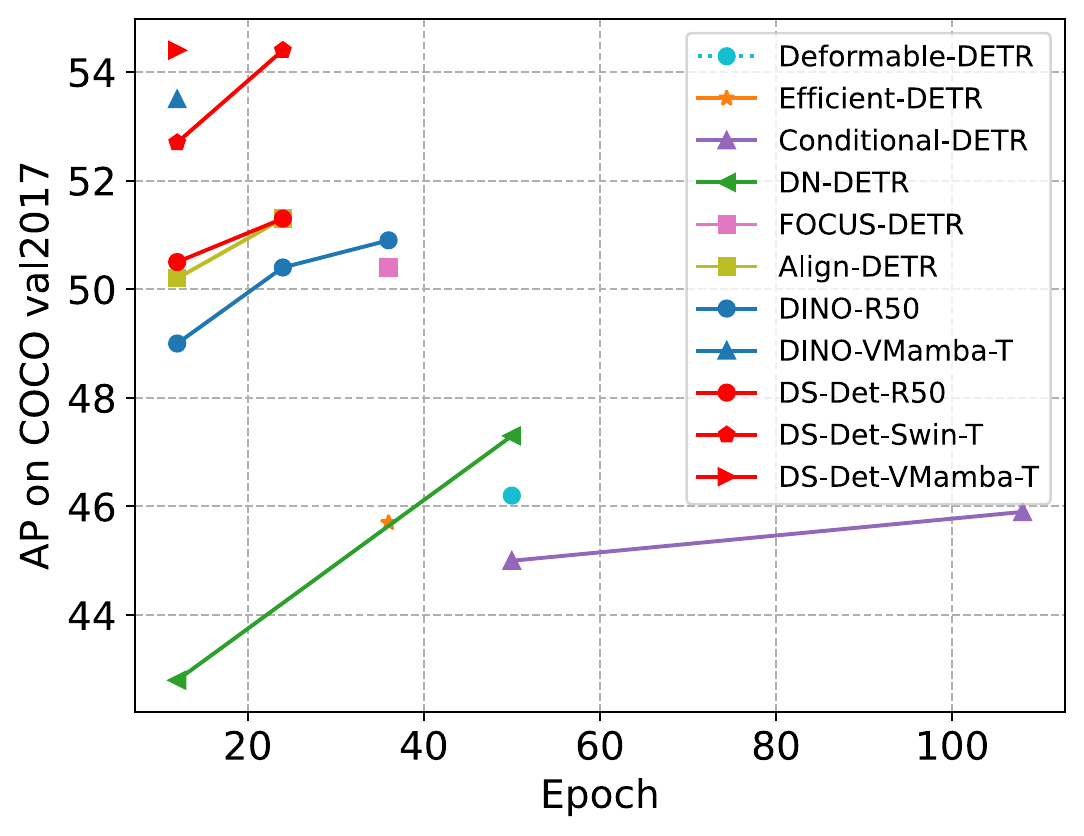}
            \subcaption{AP-Epoch}
		\end{minipage}
  } 
 {
		\begin{minipage}[t]{0.24\linewidth}
			\centering
			\includegraphics[width=0.99\linewidth]{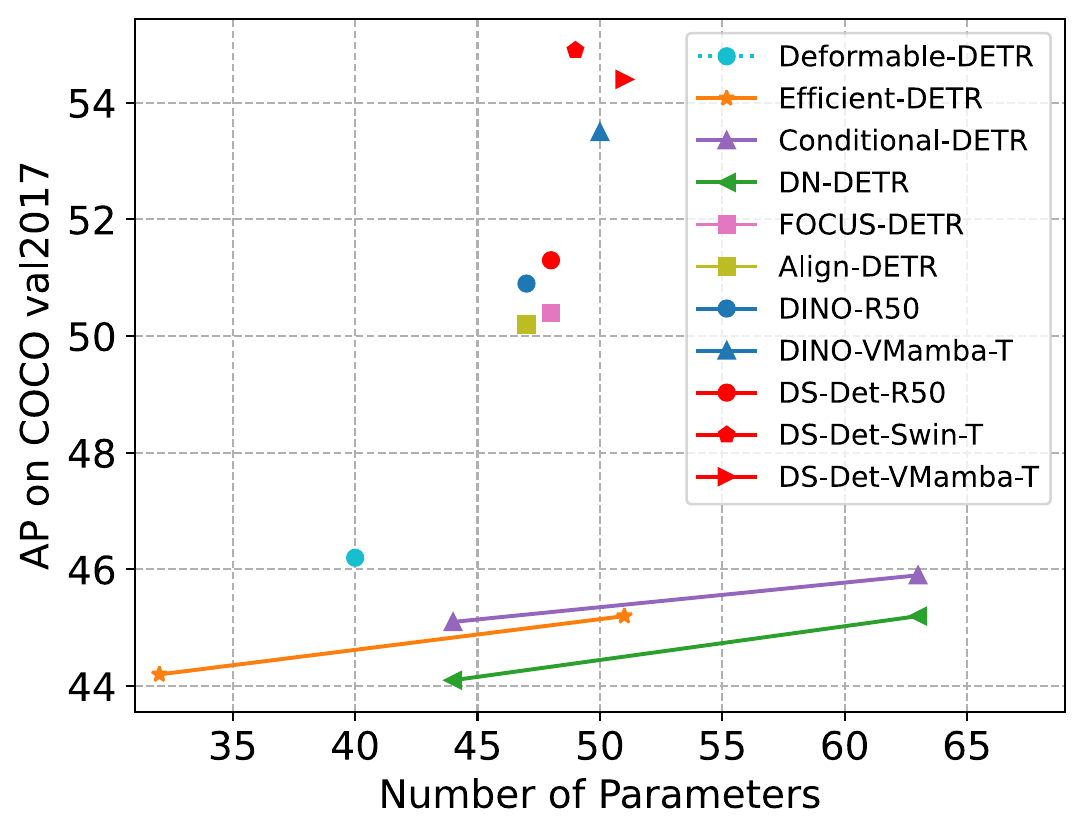}
            \subcaption{AP-Params}
		\end{minipage} 
  }   
{
		\begin{minipage}[t]{0.24\linewidth}
			\centering
			\includegraphics[width=0.99\linewidth]{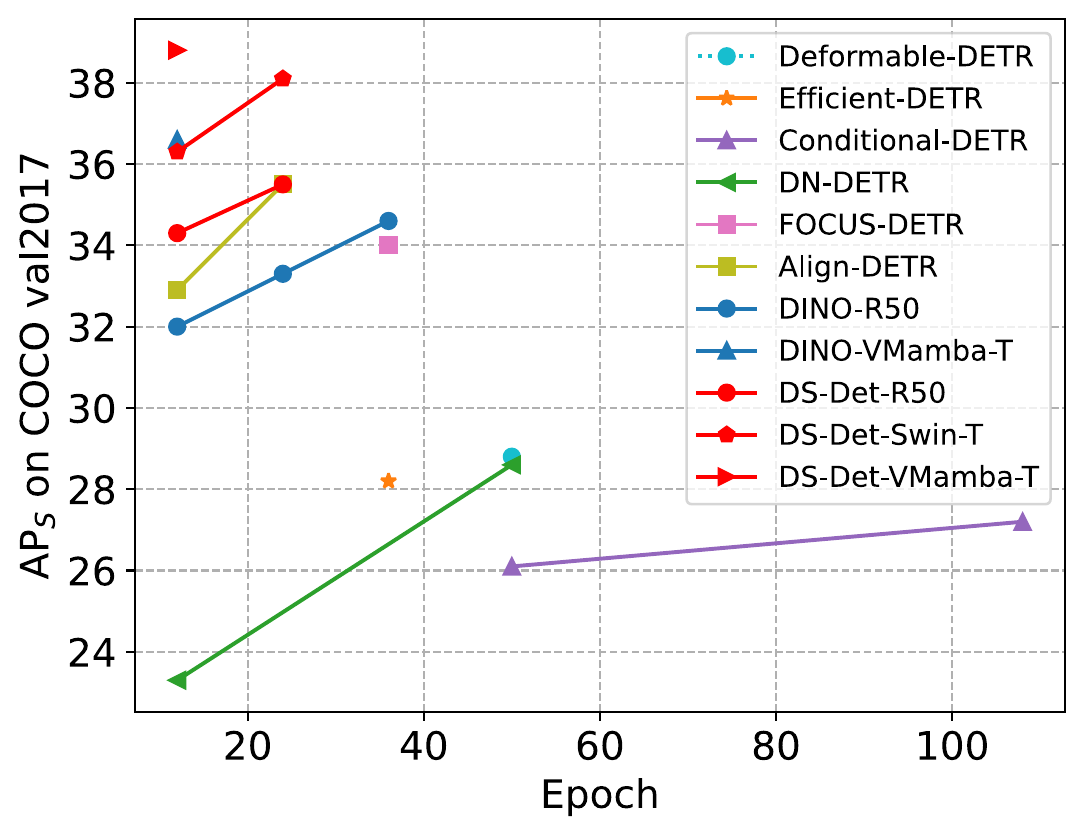}
            \subcaption{AP$_S$-Epoch}
		\end{minipage}
  }
  {
		\begin{minipage}[t]{0.24\linewidth}
			\centering
			\includegraphics[width=0.99\linewidth]{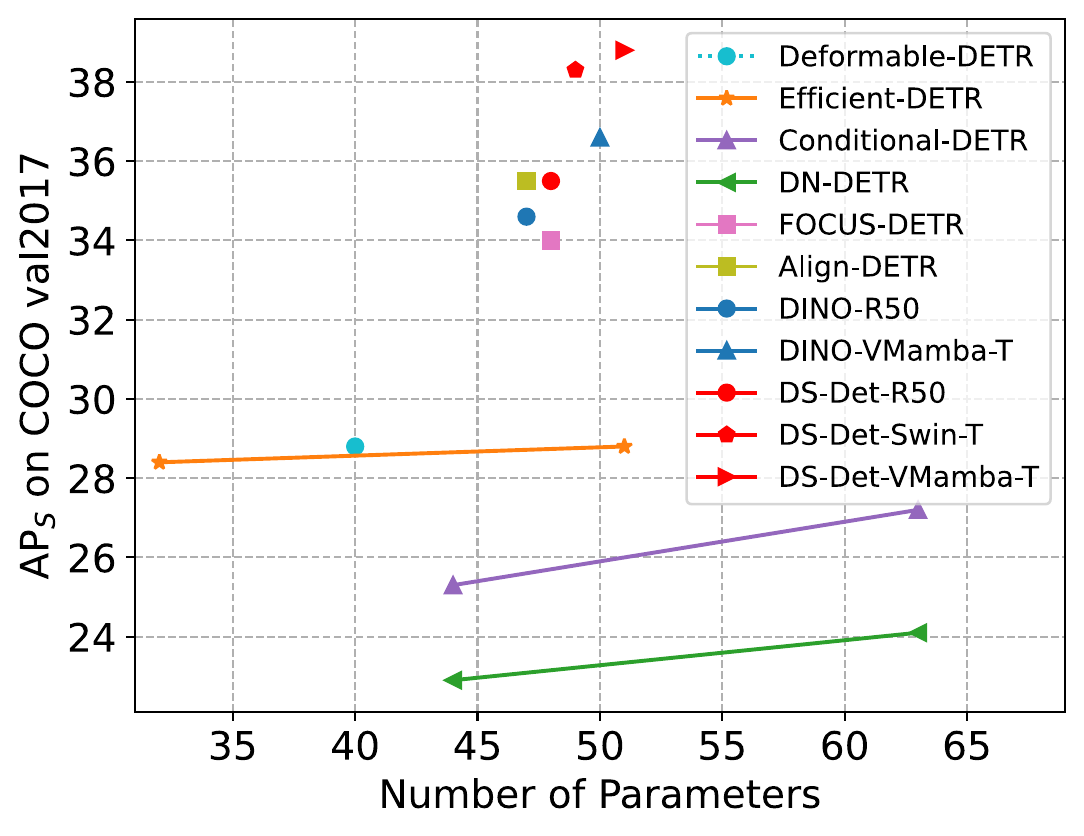}
            \subcaption{AP$_S$-Params}
		\end{minipage}
  }  
     \caption{Comparison of transformer-based models of AP and AP$_S$ w.r.t the different number of parameters and training epochs on \texttt{val2017} of COCO.
      }
    \label{fig:ap}
\end{figure*}

Figure~\ref{fig:ap} compares the performance of different transformer-based detectors on the standard detection benchmarks of the COCO dataset. The results indicate that our DS-Det model has significant advantages in terms of training efficiency (AP-Epoch, AP$_S$-Epoch), fewer parameters (AP-Params, AP$_S$-Params) on the performance of general object detection and small object detection. 


\textbf{Potential effectiveness on the downstream Open-ended detection tasks.} The primary limitation of \textit{fixed-query} detectors is their requirement to predict a \textbf{large} and \textbf{fixed} number of detection results for both \textbf{sparse} and \textbf{dense} scenes. On the one hand, this results in a considerable amount of \textit{unnecessary and redundant computation}. On the other hand, when these fixed-query detectors are applied to more challenging downstream tasks, such as Open-Ended Detection~\cite{lin2024generative}, these redundant queries would be further fed into the \textbf{large language model} (LLM) to generate object names directly, which will significantly increase the overall computational complexity and cost.

To demonstrate this, we conducted a simple test using the public official code of GenerateU~\cite{lin2024generative}. We
evaluated the computational complexity of the generative large language model used in GenerateU. With 900 queries of the detector fed into the LLM, the complexity is \textbf{10,139.64 GFLOPs}. In contrast, with 200 queries, the complexity dropped to \textbf{2,253.45 GFLOPs}, which indicates that reducing the number of queries can significantly decrease the computational load of the LLM model for the open-ended detection multi-modal task.

Based on the analysis, our flexible model, which adapts the number of queries based on the image itself, \textit{holds significant potential for open-ended object detection task}. 
This adaption characteristic enables reduced queries to feed into a generative large language model, thereby decreasing the workload for subsequent category generation processing, which highlights its potential applications and importance for future research in the vision community.

\section{Discussion}
\label{app:discussion}

\subsection{The Deduplication Roles of Self-Attention (SA) and NMS}

Removing duplicate detection boxes is a crucial step for reducing false positive samples in detection systems. We categorize existing methods of deduplication into two types based on their principles: \textbf{box-based} and \textbf{class-based}. Box-based methods, such as NMS, work by \textit{comparing the overlapping (IoU) between predicted boxes}. While straightforward, they are sensitive to the IoU threshold and struggle with overlapping objects.
In contrast, class-based methods utilize self-attention to compare features between queries, influencing classification scores to achieve deduplication, which results in \textit{lower scores for redundant boxes}.
This approach can mitigate the issue of overlapping objects and is commonly employed in DETR-like models.

Intuitively, combining both \textit{box-based} and \textit{class-based} methods may further enhance the model's overall performance. DETA~\cite{ouyang2022nms} and DDQ-DETR~\cite{zhang2023dense} both use NMS to eliminate duplicates in their models.
Compared to DETA and DDQ-DETR, our approach \textbf{addresses different problems} (in motivation and objective), \textbf{introduces different solutions to issues} (in query generation, decoder architecture, and loss function), and \textbf{achieves comparable results} in performance.

DETA is a \textit{box-based} method for deduplication. Specifically, DETA is designed to investigate the impact of \textit{one-to-many assignment-based training} on enhancing the training efficiency for DETR models, which \textbf{differs} to our motivation and objective.
It achieves this by employing one-to-many \textbf{IoU assignments} in conjunction with NMS method, which is applied during both query selection and final prediction post-processing. This study shows that the \textit{one-to-many IoU assignments}, \textit{combined with NMS}, effectively improve training efficiency.

DDQ-DETR combines both \textit{box-based} and \textit{class-based} deduplication methods in different stages. It specifically explores how \textit{query distinctness} affects the model's optimization process and accuracy, which also \textbf{differs} to our motivation and objective.
DDQ-DETR introduces the Distinct Query Selection (DQS) module, which uses training-unaware NMS to filter dense queries into distinct queries (\textit{box-based method}). Then, DDQ-DETR further applies Hungarian matching (\textit{class-based method}), considering both bounding box scores and class scores, to generate final one-to-one detection results.
Additionally, DDQ-DETR also uses one-to-many label assignments and incorporates an \textit{auxiliary head} along with Auxiliary Loss for Dense Queries to maintain training efficiency (however, this kind of mixing label assignments on same decoder weights still introduces the issue of ``query ambiguity''). Overall, this approach highlights that both sparse and dense queries in end-to-end detection are problematic. By explicitly combining box-based and class-based methods, DDQ-DETR ultimately enhances model accuracy.

Unlike these two methods, our approach focuses on transforming a \textit{fixed-query} detector into a \textit{flexible-query} detector and addressing the ``query ambiguity'' issue of DETR-like models.
We deeply explore the role of SA (which has NOT been examined in DDQ-DETR) to address challenges related to the existing decoder structure.
In contrast to DETA and DDQ-DETR, \textit{our object is not to investigate how to enhance model accuracy through NMS}.
Furthermore, we observe that the existing connection between CA and SA can result in the ``\textbf{recurrent opposing interaction}'' problem. To tackle these challenges, we developed a more effective decoder structure (ADD) and implemented \textit{decoupled} one-to-one and one-to-many label assignments. 
This design significantly alleviates the ``query ambiguity'' issue. Notably, DETA and DDQ-DETR share a similar decoder structure with existing methods, which highlights the contribution of our novel ADD decoder.
Compared to previous class-based deduplication methods, DS-Det further \textit{optimizes the decoder structure, enabling flexible detection while minimizing the ``query ambiguity'' associated with one-to-many assignments}. 


\textbf{Deduplication Part (DP) vs. NMS.} Our proposed DP is an integral part of the decoder and not a \textit{post-processing} algorithm. Although we designed DP to eliminate duplicates, it is distinct from NMS and cannot be directly replaced by it. Our model's decoder, similar to those in existing DETR-like models, comprises six layers: 4 LBP layers and 2 DP layers, connected in series. As analyzed in DETR~\cite{carion2020end}, the decoder primarily reasons about the relations of queries and image context to generate detections, which is \textit{necessary} for the detector. In Table~\ref{tab:ablation_blp_dp}, we conducted ablations on the BLP and DP layers. The results indicate that reducing the number of DP layers (for example, using only one DP layer) is harmful to the performance, underscoring the importance of the DP layer.

In Table \textcolor{purple}{2} of the main paper, we present a \textit{flexible-query} model that removes all SA (approximating the removal of the DP module), while employing the NMS to eliminate duplications. This model achieved only 47.5\% AP on COCO, which is significantly lower than the 50.5\% AP achieved with our model with DP, clearly highlighting the importance and irreplaceability of the DP module.

On the other hand, NMS can only perform deduplication on the detection results that have already been obtained, and it is sensitive to hyper-parameters; it cannot generate detection results directly, making it unsuitable as a replacement for the decoder. The DETR model was originally designed to simplify the detection process in an \textbf{end-to-end} pipeline, removing post-processing steps like NMS and improving model robustness. \textit{Our method aligns well with this goal}, enabling the model to function effectively without relying on NMS, but not to pull NMS back again.

The original DETR~\cite{carion2020end} also notes that the Feed Forward Network (FFN) in the decoder has a significant impact on model accuracy. We have made a test on the FFN of the DINO model to reduce computational complexity by decreasing the hidden layer dimension in the FFN from 2048 to 768. However, this change resulted in a \textbf{1.6\%} decrease in AP, underscoring the critical role of the decoder network layers in maintaining performance. This further underscores the importance of the decoder.

\subsection{The Differences of Decoupled Matching with Existing Methods}

One-to-one matching and one-to-many matching label assignments have been demonstrated in several studies (such as H-DETR~\cite{jia2023detrs}, Align-DETR~\cite{cai2023align}, DAC-DETR~\cite{hu2024dac}, etc) to enhance the training convergence by increasing the number of positive query samples.  
Unlike simply mixing one-to-one and one-to-many matching label assignments, our method features a decoupled matching process that constructs a \textbf{new decoder} ADD. This design is based on experiments and an analysis of the roles of SA and CA and addresses the need to transition from a fixed-query to a flexible-query detector.

In this paper, we address the \textbf{query ambiguity} that arises from the mixed usage of one-to-one and one-to-many matching. We tackle this issue at the label assignment level by designing the ADD decoding structure with one-to-many label assignment in BLP and one-to-one label assignment in DP, which \textit{decouples} the two types of matching while enabling each to \textit{fulfill} its role effectively. At the same time, we explore \textit{another dimension} of one-to-one and one-to-many processes (\textit{beyond the label assignment level}), specifically the opposing effects of CA and SA. By effectively leveraging the CA and SA in the decoder structure, we have alleviated the ``\textbf{recurrent opposing interactions}'' issue associated with both CA and SA. This is illustrated in Appendix Table~\ref{tab:ablation_blp_dp} (the ablation on BLP and DP layers) and Table~\ref{tab:ablation_reverse} (the ablation on the order of CA and SA), which highlight our ``decoupled matching'' impact on accuracy.


\textbf{Differences between H-DETR, MS-DETR, and DS-Det on Decoupled Matching.} Specifically, our work differs from H-DETR~\cite{jia2023detrs} and MS-DETR~\cite{zhao2024ms-detr} in four main aspects:

\textbf{(1) Problem Addressed:} The primary goal of both H-DETR and MS-DETR is to enhance the model's \textit{training efficiency} to speed up the training convergence. In contrast, our DS-Det model focuses on resolving ``\textbf{query ambiguity}'' caused by mixed label assignments (i.e., combining one-to-one and one-to-many) and addressing the ``\textbf{recurrent opposing interaction}'' issues that arise from the interaction between CA and SA.

(2) \textbf{Design of Decoder Structure:} H-DETR uses \textit{additional} branches to learn one-to-many assignments, which significantly increases the training cost. 
The MS-DETR shares a \textit{similar} decoder structure as DINO and introduces \textit{additional heads} (box and class predictors) for one-to-many supervised to further enhance the training efficiency.
Different from these two methods, our approach employs a \textbf{single} branch and implements an efficient end-to-end structure through disentangled attention learning, with BLP for locating boxes and DP for refining and de-duplicating detections. Our approach effectively alleviates the ``query ambiguity'' from mixing label assignments to \textit{decoupled label assignments}, incorporating with the optimized decoder structure by leveraging the unique characteristics of CA and SA to address the ``\textit{recurrent opposing interaction}'' problem, \textit{not only ensuring faster convergence, reducing the complexity, but also further mitigating the ``query ambiguity'' at the same time}.

(3) \textbf{Detection Capability:} H-DETR and MS-DETR are both limited to detecting a \textbf{fixed} number of objects, whereas our decoder is designed to detect an \textbf{flexible} number of objects, which is beneficial for many applications, such as sparse/dense/open-ended detection tasks.

(4) \textbf{Detection Accuracy:} Compared to H-DETR, as shown in Table \textcolor{purple}{3} of the main paper, with the same backbone (Swin-T), our model achieves higher performance, increasing by \textbf{+2.1\%} AP (52.7\% vs. 50.6\%) and \textbf{+2.9\%} AP$_S$ (36.3\% vs. 33.4\%) (Table \textcolor{purple}{3} of the main paper). Compared to MS-DETR, with the same backbone (ResNet50), our model also achieves higher performance, increasing by \textbf{+0.2\%} AP (50.5\% vs. 50.3\%) and \textbf{+1.6\%} AP$_S$ (34.3\% vs. 32.7\%) than MS-DETR, further confirming DS-Det's effectiveness. 
As mentioned in MS-DETR, we also believe that the one-to-many supervision using additional head modules in MS-DETR is a \textit{complementary approach} to our model, which could \textit{potentially} further enhance the training efficiency and accuracy of our method. Additionally, our DS-Det significantly surpasses another flexible model of Diffusion-Det~\cite{chen2023diffusiondet} model by \textbf{+4.5\%} AP (51.3\% in 24 epochs of DS-Det vs. 46.8\% in 60 epochs of Diffusion-Det), further confirming its effectiveness.

\subsection{The ADD Framework for Addressing the Query Ambiguity Issue}
\label{app:pq_remove}

In the main paper, we have discussed that ``query ambiguity'' arises from two main reasons: one-to-one and one-to-many label assignments with shared decoder weights, and the opposing roles of CA and SA with ``\textit{recurrent opposing interactions}'' operation.
To address this, we designed a unified ADD framework that effectively decouples mixing label assignments and explicitly removes the ``recurrent opposing interactions'' operations. 

Specifically, the BLP module only contains the CA module and employs the one-to-many supervision mechanism, while the DP module incorporates multiple SA and employs the one-to-one matching label assignment mechanism. These designs effectively eliminate the sources of query ambiguity from the outset.

When connecting the BLP and DP modules, the conflict between the two label assignments still exists, due to the query being sequentially updated by the BLP and DP modules. To address this, we designed a simple yet effective method for Stopping Gradient back-propagation of Query (SGQ) from DP to BLP, which helps mitigate conflicts between the two modules. As shown in Table \textcolor{purple}{9}, the absence of the SGQ approach resulted in a \textbf{2.0\%} AP drop in the model's performance, highlighting the significance of our approach in alleviating query ambiguity and further validating ADD's effectiveness.

\subsection{The Computational Complexity Analysis}

In this section, we analysis the model's computational complexity in four aspects:

(1) For the \textit{computational complexity}, we have included analysis in Sec.~\ref{sec:complexity_analysis_1}, indicating that our model can effectively reduce the number of queries, and the corresponding
FLOPs are also reduced, as illustrated in Fig.~\ref{fig:flops}. 

(2) For the \textit{efficiency} comparison, we have conducted experiments on both COCO and WiderPerson datasets to show the relations between the number of objects, adaptive query numbers, and the corresponding performance (Sec.~\ref{sec:experimental_test}). The results (in Table~\ref{tab:relation}, Fig.~\ref{fig:query_acc}) indicate that our model obtained overall higher performance across all subsets of COCO than DINO. Especially for the subset of 1-5, our model only uses \textbf{7.25\%} of the queries (65.22 vs 900) while achieving higher performance by \textbf{+0.9\%} AP. For the more challenging dataset of WiderPerson, our model obtains overall higher performance under both sparse and dense scenes (in Table~\ref{tab:widerperson_30}, Fig.~\ref{fig:wid_sts}). These results efficiently demonstrate the effectiveness of our method.

(3) For the inference time, we conducted \textit{additional} comprehensive tests on inference speed (in Sec. \textcolor{purple}{4.3} of the main paper), varying the number of queries from 10 to 1800. The results indicate that, with
the same number of queries as DINO (900 queries), our ADD
decoder framework has improved the inference speed of the
model’s decoder by \textbf{+34.8\%} over DINO.

(4) In terms of \textit{inference} memory usage, our model is similar to DINO. During inference, modules like CA and SA generate \textit{intermediate variables}, causing dynamic changes in GPU memory. We test and record the \textbf{maximum memory allocation} (batch size = 1): 912.3 MB for DINO (900 queries) and 914.9 MB for DS-Det (900 queries), indicating only a small difference between the two models. This slight variation may be due to differences in the implementation of decoder structures, such as the intermediate variables within the code. Since the adaptive number of queries learned by the model is limited to a small range, the differences in memory usage are minimal. 
It is noteworthy that the significance of the model's flexible number of queries, derived from the image itself, lies in \textit{its ability to detect a flexible number of objects while reducing computational load}, as illustrated in Sec.~\ref{sec:complexity_analysis_1}.





\textbf{How does the model scale with increasing image complexity and object density?}
In Sec.~\ref{sec:experimental_test}, we have tested the proportional relation between the number of objects and the number of queries. The results show that as the number of objects in an image increases, the rate at which the number of queries increases slows down. To better illustrate this, we calculated the ratio of the model's queries to the number of objects in the image, presented in Table~\ref{tab:relation}. From the table, it can be observed that as the number of objects increases, the growth rate of queries \textit{slows down} and gradually declines, falling from 25.08 to 16.37. Concurrently, our model demonstrates a \textit{growing advantage} over DINO as the number of objects increases across the subsets (31–35, 36–40, 41–45, 46+), as illustrated in Fig.~\ref{fig:query_acc}. Notably, the subset of 1-5 objects accounts for 55.9\% of the images in the val2017 set of COCO. For this subset, our model utilizes only \textbf{7.25\%} of the queries (65.22 vs. 900) while achieving a higher performance with a \textbf{+0.9\%} AP, suggesting that the queries selected via FLET module are more effective. 

In terms of complexity, we have discussed the changes in FLOPs with varying queries in Sec.~\ref{sec:complexity_analysis_1}. As shown in Fig.~\ref{fig:query_acc}, the reduction in the number of queries (cyan dashed line) effectively lowers the model's FLOPs (purple solid line). These experiments underscore the effectiveness of our method's adaptive characteristics in object detection.

\textbf{Trade-offs between performance and computational cost in DS-Det.} For the model with a flexible number of queries, computational complexity and performance are related to the number of objects in the test images. In the Sec.~\ref{sec:complexity_analysis_1}, we have conducted ablation studies on the classification score $S$ using the COCO dataset, exploring how the number of queries and FLOPs change with variations in the classification score \textit{S}. From Table~\ref{tab:ablation_threshold} and Fig.~\ref{fig:flops}, we can observe that a lower threshold \textit{S} leads to a higher number of queries, which improves the model's accuracy but also increases its computational complexity. When \textit{S} ranges from 0.01 to 0.05, the model's performance varies in a small range of
50.2\% $\sim$ 50.6\% AP, indicating good robustness. To balance the trade-offs between performance and computational cost, we set $S=0.02$ for the configuration used in other experiments in the main paper. With this configuration, our model achieves higher performance (50.5\% vs. 49.0\% in AP) while maintaining smaller computational complexity (273G vs. 279G) and faster decoder inference speed (9.2 ms vs. 14.1 ms) than DINO.

\subsection{The Contributions of FLET module}

The decoder query plays a crucial role in reasoning about the relations of the object and the global image context to output the detections, connecting the components of the decoder and greatly impacting the performance. As discussed in Sec. \textcolor{purple}{3.1}, the process of transforming a fixed-query detector into a flexible-query detector is influenced by the \textit{decoder architecture, the query components and roles, and the use of the self-attention module}.
Specifically, our FLET module is novel in three key aspects:

(1) \textbf{New Query Paradigm}: FLET module introduces a new query type that alters the existing query composition in DETR models (from content and positional queries to Single-Query). This approach effectively leverages encoder token information and reduces the need for random embeddings for query initialization. Most existing methods directly follow the query design of DETR, where the positional query is originally adopted to \textit{increase the difference between queries to produce different results} (as outlined in Sec.3.2 of DETR~\cite{carion2020end}). However, our experiment and analysis indicate that the \textit{positional query is unnecessary}, which is an important insight for future research on both the fixed-query and flexible-query detectors.

(2) \textbf{Flexible Queries}: FLET module switches from a fixed, predefined number of queries to a flexible number of queries through the encoder-to-query mechanism. As the model training progresses, this method gradually enables the classification head to distinguish between positive and negative samples among all encoder tokens, obtaining a global solution.

(3) \textbf{Addressing Training GPU Memory Limitation for Using SA module}: FLET module effectively addresses training GPU memory limitations, allowing for training the model with the essential self-attention (SA) module within the decoder architecture. In contrast, directly switching from fixed-query to flexible-query without SA would significantly degrade performance (e.g., a \textbf{4.9\%} AP drop, as shown in Table~\textcolor{purple}{2} of the main paper).

\subsection{Advantages of the DS-Det Model}

The primary limitation of \textit{fixed-query} detectors is their requirement to predict a \textbf{large} and \textbf{fixed} number of detection results for both \textbf{sparse} and \textbf{dense} scenes. On one hand, this results in a considerable amount of \textit{unnecessary and redundant computation}. On the other hand, when these fixed-query detectors are applied to more challenging downstream tasks, such as Open-Ended Detection~\cite{lin2024generative}, these redundant queries would be further fed into the \textbf{large language model} (LLM) to generate object names directly, which will significantly increase the overall computational complexity and cost, as illustrated in Sec.~\textcolor{purple}{4.3} of the main paper.

In contrast, DS-Det \textit{adaptively} eliminates redundant queries at \textit{an early stage} in transformer detector. This leads to a \textbf{more efficient, cost-effective, accurate, and flexible approach}, as demonstrated in the following four aspects:

(1) \textbf{High-rate of effective query utilization, better cost-efficiency, and higher performance.} In \textit{sparse scenarios}, DS-Det achieves comparable or even higher accuracy with only a small number of queries, while simultaneously reducing the computational load of the decoder. The transformer-decoder is primarily composed of layers of cross-attention (CA) and self-attention (SA). CA has a complexity of \textbf{$o(NKC^2)$} (using deformable attention~\cite{zhu2020deformable}, where $N$ is the number of queries, $K$ is the number of sample points, and $C$ denotes the number of channels), while SA has a complexity of \textbf{$o(N^2)$}. As the number of queries $N$ decreases, the computational load of the CA and SA reduces at a linear and quadratic rate, respectively. This makes the approach \textit{highly cost-effective} and results in \textit{lower power consumption} during deployment, particularly in \textit{edge devices}.

In many scenarios, objects within the image are often sparse. For instance, in the val2017 of the COCO dataset, \textbf{55\%} of images contain between 1 and 5 objects (as shown in Sec.~\ref{sec:experimental_test}, Table~\ref{tab:relation}). In this context, we achieved an accuracy that is \textbf{+0.9\%} AP higher than the DINO model while using only \textbf{7.25\%} of the queries (65.22 vs. 900), clearly demonstrating the effectiveness of our query with high-rate utilization.

(2) In dense scenarios, our flexible model also achieves higher accuracy with fewer queries than DINO. This robust advantage becomes even more pronounced as the number of objects increases, as illustrated in Fig.~\ref{fig:query_acc} and Table~\ref{tab:relation} (specifically in the subsets of $31\sim35$, $36\sim40$, $41\sim45$, $46+$).

(3) For more complex \textit{multi-modal} detection tasks, such as open-ended detection~\cite{lin2024generative}, decoder queries are inputted into \textbf{large language models} to directly generate corresponding object names without additional vocabulary priors. This process is highly \textbf{complexity-sensitive} to the number of \textit{queries} (for instance, the LLM model shows 10,139.64 GFLOPs with 900 queries and 2,253.45 GFLOPs with 200 queries~\cite{lin2024generative}). However, fixed-query approaches that using a large fixed number of queries significantly increase the computational load. In contrast, our flexible method reduces this redundancy by eliminating unnecessary queries at an early stage, which is highly significant and holds great potential for these multi-modal tasks.

(4) Our flexible method offers greater \textbf{flexibility} through its inherent adaptive characteristics. For example, once the model is trained, the number of queries can be adaptively adjusted to suit \textit{various scenarios} without the need for retraining, which is still required for fixed-query detectors.

Furthermore, our model has demonstrated superior performance on the COCO dataset, as well as on the more challenging WiderPerson and CrowdHuman datasets compared to DINO. This further indicates its \textit{robustness} across a variety of scenarios.

In addition, we also introduce a novel decoding framework, ADD, which effectively tackles the ``query ambiguity'' issue caused by mixing label assignments of one-to-one and one-to-many, as well as the ``ROT'' problem. The effectiveness of ADD framework is demonstrated by the experimental results in Table~\ref{tab:ablation_blp_dp} and Table~\textcolor{purple}{9} of the main paper. It \textit{enhances} performance while \textbf{simplifying} the decoder architecture by using \textit{fewer} SA layers, a single adaptive query type, and \textit{eliminating the need for additional branches or multi-head prediction modules}.

Finally, our ADD framework significantly improves the speed of the decoder by \textbf{+34.8\%} compared to DINO (as detailed in Sec.~\textcolor{purple}{4.3} of the main paper). Although the overall speed advantage may not be substantial due to the multiple components of transformer-based detectors——where the slow inference speed is a common bottleneck for transformer detectors, especially when compared to faster CNN-based detectors like YOLO~\cite{khanam2024yolov11})—— it offers a promising solution for enhancing the speed of transformer-based detectors. For instance, by integrating the YOLO \textit{backbone}, the Sparse-DETR~\cite{roh2021sparse} \textit{encoder}, and our ADD \textit{decoder}, we may have great potential to develop a \textit{high-speed, low-cost, high-performance, flexible} transformer detector, which deserves further exploration in future work.

\section{Visualization of Flexible Queries on \texttt{val2017} of COCO}
\label{app:visualizations}

This paper proposes a novel transformer-based \textit{flexible} detector that can predict a flexible number of objects for different input images. By adaptively generating queries from encoder tokens, our model significantly improves query efficiency by reducing redundant queries and decreasing the computational cost. 

\begin{figure*}[!htbp]
  \centering
   \includegraphics[width=0.85\linewidth]{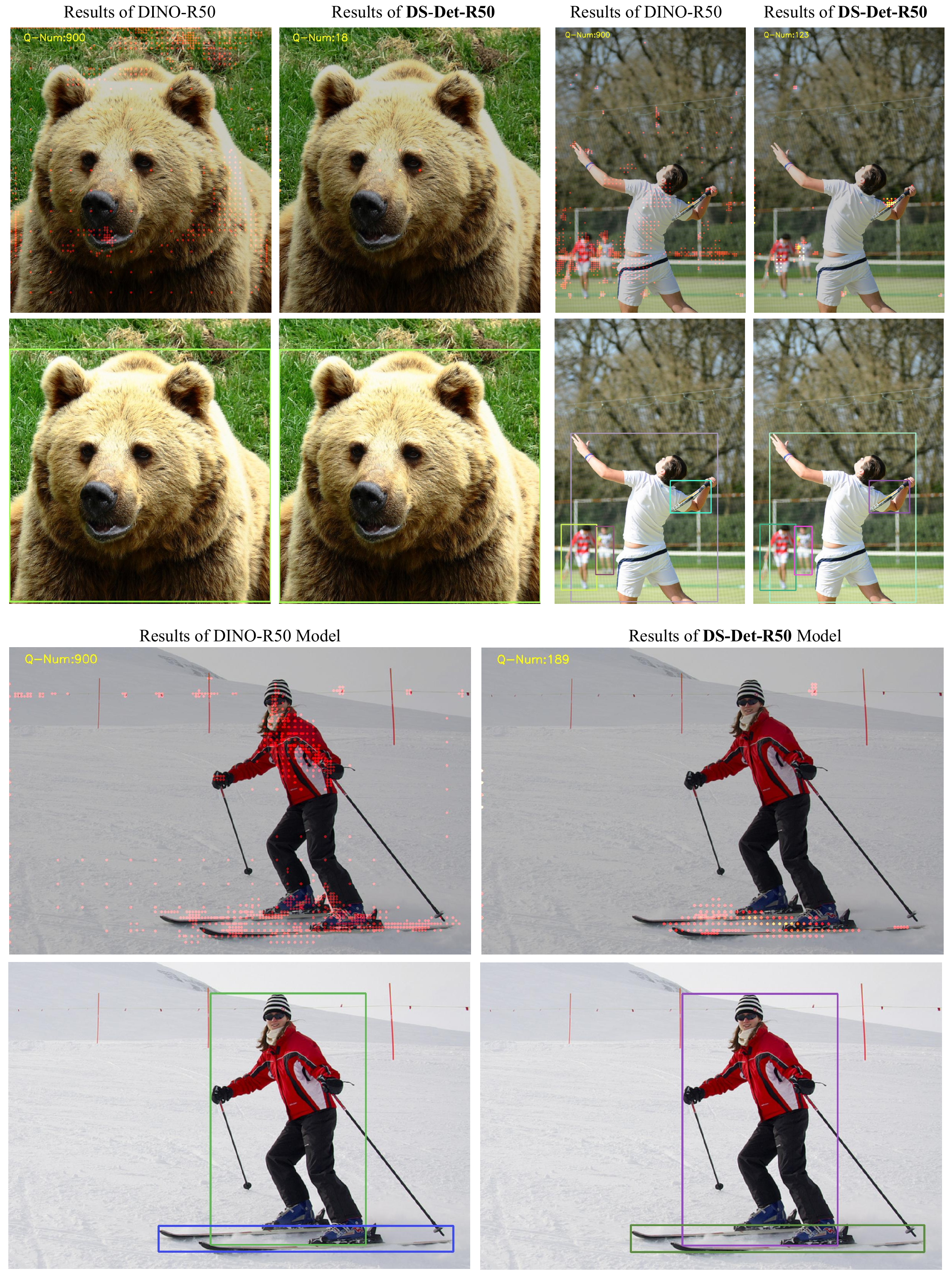}
   \caption{Visualization of queries in simple scenarios. The number of queries (Q-Num) selected for each image is on the top left corner of the corresponding image.}
   \label{fig:vis1}
\end{figure*}

\begin{figure*}[!htbp]
  \centering
   \includegraphics[width=0.80\linewidth]{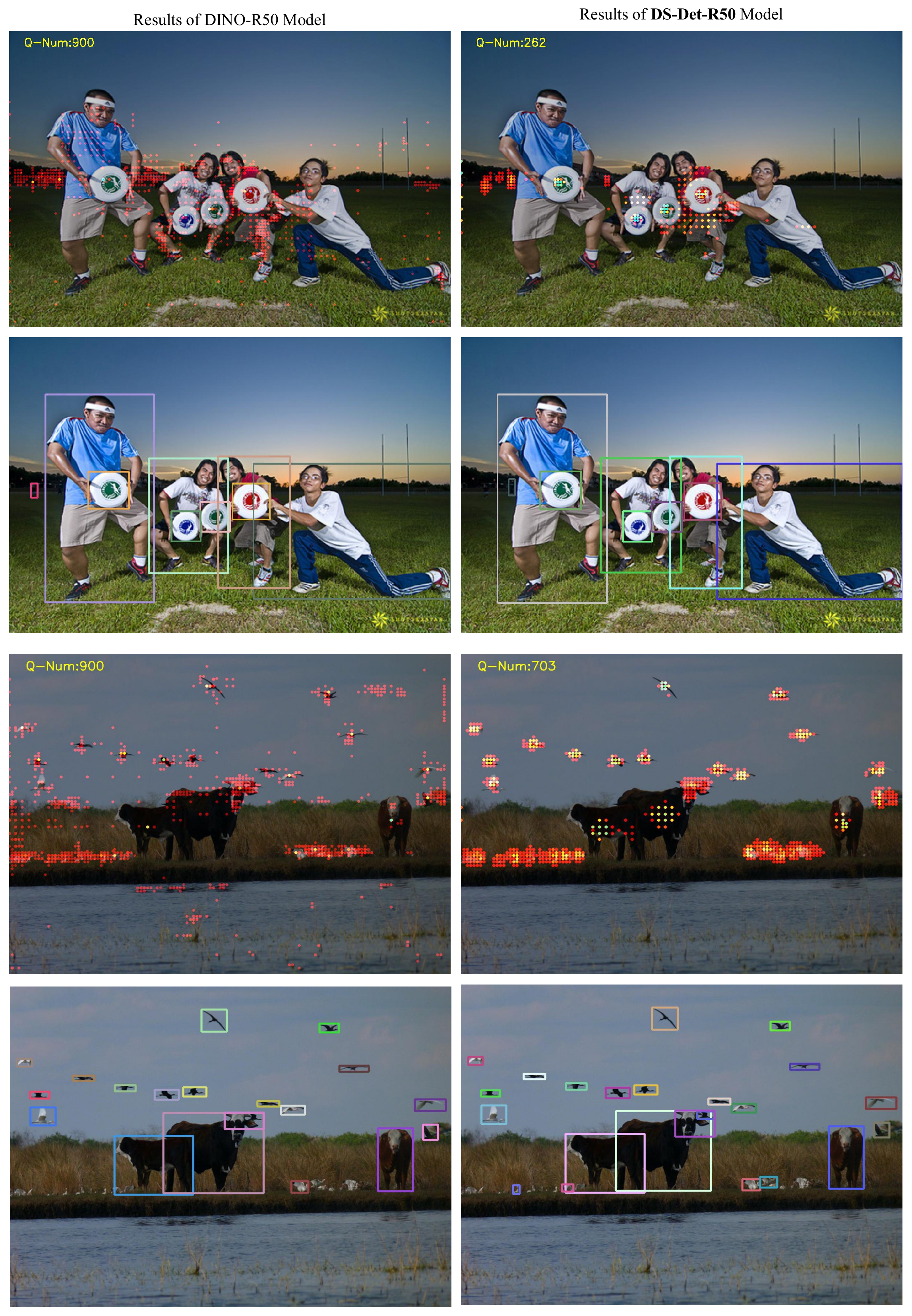}
   \caption{Visualization of queries in complex scenarios.}
   \label{fig:vis2}
\end{figure*}

\begin{figure*}[!htbp]
  \centering
   \includegraphics[width=0.80\linewidth]{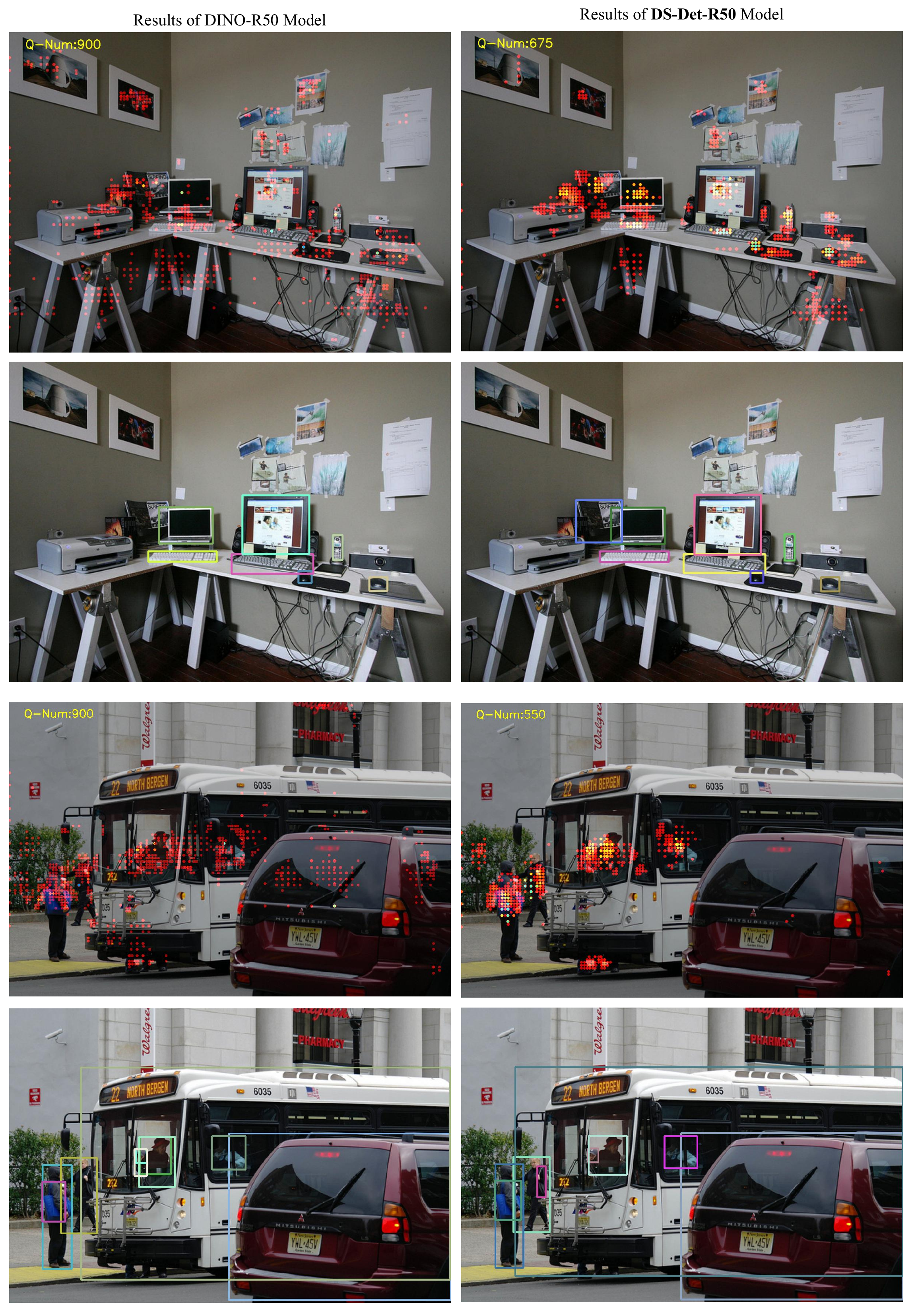}
   \caption{Visualization of queries in complex scenarios.}
   \label{fig:vis3}
\end{figure*}

To visually demonstrate this, we compare the test results of our DS-Det-ResNet50 (12 epochs) model and the DINO-ResNet50 (12 epochs) model on the \texttt{val2017} of COCO dataset in Fig.~\ref{fig:vis1}, Fig.~\ref{fig:vis2}, and Fig.~\ref{fig:vis3}.
To ensure clear visualization, the queries are represented using solid circle points with a radius of 3, where the color of the circles indicates the different classification confidence scores. The detection bounding boxes are displayed using random colors to differentiate the different object instances.

In simple scenarios, such as the bear's detection in Fig.~\ref{fig:vis1}, DS-Det-ResNet50 only used 2\% of the queries compared to DINO-ResNet50, yet achieved accurate detection results. 
In the relatively complex scenarios with more objects in Fig.~\ref{fig:vis2} and Fig.~\ref{fig:vis3}, DS-Det-ResNet50 similarly adapted and selected fewer queries while achieving more precise detection results. 
This demonstrates that the DS-Det model can adaptively select the number of queries based on different image inputs, thereby enabling the ``flexible predictions''. This approach also reduces the number of redundant queries, effectively improving the model's performance and efficiency.

